\documentclass{article}

\usepackage{microtype}
\usepackage{graphicx}
\usepackage{subcaption}
\usepackage{booktabs}
\usepackage{xspace}
\usepackage{xcolor}
\usepackage{enumitem}
\usepackage{tabularx}

\newcommand{\lex}[1]{\colorbox{yellow!35}{#1}}
\newcommand{\orth}[1]{\colorbox{blue!20}{#1}}
\newcommand{\morph}[1]{\colorbox{green!25}{#1}}

\usepackage[colorlinks=true, allcolors=blue!60!black, hyperfootnotes=false]{hyperref}

\newcommand{\methodName}{DialectLLM\xspace}
\newcommand{\benchmarkName}{DialectLLM-Bench\xspace}

\PassOptionsToPackage{numbers,sort&compress}{natbib}
\usepackage[main, final]{neurips_2026}

\usepackage{algorithm}
\usepackage{algorithmic}
\usepackage{amsmath}

\usepackage{amssymb}
\usepackage{mathtools}
\usepackage{amsthm}

\usepackage[most]{tcolorbox}
\usepackage{courier} % For the typewriter font style

\definecolor{judgeheader}{RGB}{0, 32, 160} % Dark Blue
\definecolor{judgeback}{RGB}{242, 242, 255} % Pale Lavender/Blue

\newtcolorbox{promptbox}[1]{
    enhanced,
    colback=judgeback,
    colframe=judgeheader,
    title={#1},
    fonttitle=\sffamily\bfseries\large,
    coltitle=white,
    arc=3mm,                % This creates the rounded corners
    drop shadow={black!40!white}, % The shadow
    boxrule=1.5pt,
    left=10pt,
    right=10pt,
    top=8pt,
    bottom=8pt,
    before skip=15pt,
    after skip=15pt,
    halign=flush left,
}

\usepackage[capitalize,noabbrev]{cleveref}

\theoremstyle{plain}

\theoremstyle{definition}

\theoremstyle{remark}

\usepackage[textsize=tiny]{todonotes}

\title{DialectLLM: A Dialect-Aware Dialog[ue] Generation Framework Beyond Standard American English}
\newcommand{\affmark}[1]{\textsuperscript{\normalfont #1}}
\author{%
\begin{tabular}{c}
Jio Oh\affmark{1,2}\thanks{Work done during internship at Amazon. Correspondance to Jio Oh: $\langle$\texttt{harryoh99@kaist.ac.kr}$\rangle$}
\quad
Paul Vicinanza\affmark{2}
\quad
Thomas Butler\affmark{2}
\\
Steven Euijong Whang\affmark{1}
\quad
Dezhi Hong\affmark{2}
\quad
Amani Namboori\affmark{2}
\\[0.5em]
{\normalfont \affmark{1}KAIST \qquad \affmark{2}Amazon}
\end{tabular}
}

\begin{document}
\maketitle

\begin{abstract}
More than 80\% of the 1.6B English speakers do not use Standard American English (SAE), yet LLMs often fail to correctly identify non-SAE dialects and generate stereotyped responses for their speakers. We introduce \textbf{DialectLLM}, the first large-scale framework for generating high-quality multi-dialectal conversational data encompassing the three pillars of written dialect---lexical (vocabulary), orthographic (spelling), and morphosyntactic (grammar) features. DialectLLM produces a dialect-parallel dialog dataset spanning nine English dialects. Partnering with native linguists, we design and validate SAE-to-dialect transformation rules, ensuring authenticity. Our approach challenges the prevailing practice of applying a single morphosyntactic feature set to both user utterances and model responses, showing that models should not reproduce up to 90\% of the grammatical features of a dialect. Human evaluation confirms data quality, with annotators preferring \methodName over prior methods in 98.8\% of pairwise comparisons for dialect naturalness. We then construct \textbf{DialectLLM-Bench}, a dialect-parallel benchmark with 50k+ dialogs, resulting in 97k+ QA pairs, and evaluate 17 LLMs on dialect identification and response generation tasks. Even frontier models achieve around 70\% accuracy, fail to reach 40\% for prominent dialects like Canadian English, and systematically misclassify non-SAE dialects as American or British. Beyond benchmarking, we show that DialectLLM data also serve as a scalable LLM post-training resource, suggesting a practical path toward dialect-aware conversational AI.

%Independent evaluations confirm data quality, with annotators preferring DialectLLM outputs, over prior methods in 98\% of pairwise comparisons for dialect naturalness. Using this pipeline, we construct the dialect-parallel \textbf{MDialBench}mark with 50k+ dialogs, resulting in 97k+ QA pairs, and evaluate 17 LLMs on dialect identification and response generation tasks. Even frontier models achieve under 70\% accuracy, fail to reach 50\% for Canadian English, and systematically misclassify non-SAE dialects as American or British. Beyond benchmarking, MDial can also serve as a scalable source for dialect-aware LLM post-training, yielding substantial gains in dialectal tasks.

%As dialect identification underpins natural language understanding, these errors risk cascading failures into downstream tasks.

%We address this gap with three contributions.

%While research on multilingual LLM capabilities has flourished,

\end{abstract}

\section{Introduction}

Large Language Models (LLMs) are increasingly deployed in conversational AI, yet their training data skew heavily toward Standard American English (SAE). The limited dialect-specific data available are often narrow in domain and exaggerated in form, causing models to learn stereotypical and potentially offensive patterns~\cite{fleisig-etal-2024-linguistic,deas-etal-2025-data}. These deficiencies persist beyond post-training: models exhibit degraded performance for non-SAE users across classification, question-answering, and reasoning tasks~\cite{multivalue,held-etal-2023-tada, liu-etal-2023-dada, xiao-etal-2023-tasklora, holt-etal-2024-perceptions,dialectbench, lin-etal-2025-assessing, gupta-etal-2025-endive, lee2025transenv}. More troublingly, LLMs have been shown to display explicit bias against users of non-standard dialects~\cite{hofmann2024ai, smith2024standard,fleisig-etal-2024-linguistic}. These deficits not only degrade user satisfaction~\cite{why_AI_dialect}, but also disadvantage dialect speakers in real-world LLM applications~\cite{lin-etal-2025-assessing}.

To narrow this gap, various dialect-specific data generation methods have been proposed. Prior approaches typically transform SAE text into a target dialect using human translation~\cite{lin-etal-2025-assessing}, rule-based heuristics~\cite{multivalue}, LLMs~\cite{gupta-etal-2025-endive}, or hybrid methods~\cite{lee2025transenv}. LLMs offer scalability, but relying on parametric knowledge alone proves insufficient: even state-of-the-art models struggle to produce natural-sounding dialect text~\cite{deas-etal-2023-evaluation, yong-etal-2023-prompting, smith2024standard}. As a result, recent works mostly rely on structured external linguistic resources such as the Electronic World Atlas of Varieties of English (eWAVE) to apply rule-guided transformations.

However, our analysis of previous work and eWAVE itself reveals three critical limitations. First, eWAVE indexes only morphosyntactic variation, omitting orthographic variations (e.g., \textit{color} vs. \textit{colour}) and lexical variations (e.g., \textit{apartment} vs. \textit{flat}), which are central to written dialect. Second, evaluation with native speakers reveals that direct application of eWAVE rules produces exaggerated and often antiquated representations of dialects. Third—and most critically—prior work conflates dialect features used by speakers with those appropriate for model generation. For example, the discourse marker \textit{like} (``Carol was, like, making lunch'') is pervasive in spoken English, including SAE, yet is implicitly understood to be inappropriate for generated written text. By treating all attested eWAVE features as suitable for model output, dialect-specific generators risk reinforcing the very stereotypes they aim to correct.

\begin{table}[t]
\caption{
Example Irish dialect transformation comparing eWAVE LLM-based generation~\cite{lee2025transenv} and \methodName. 
Highlighted spans indicate transformation type:
\lex{lexical} substitutions,
\orth{orthographic} changes,
and \morph{morphosyntactic} edits.
Note the lack of lexical or orthographic changes in prior methods, as well as the \morph{thou} transformation, which our Irish annotators flagged as archaic.
}
\centering
\small
\resizebox{0.88\linewidth}{!}{%
{\begin{tabular}{@{}p{0.31\linewidth}p{0.31\linewidth}p{0.31\linewidth}@{}}
\toprule
\multicolumn{1}{c}{\textbf{SAE baseline}} &
\multicolumn{1}{c}{\textbf{Trans-EnV (eWAVE + LLM)}} &
\multicolumn{1}{c}{\textbf{Ours (DialectLLM-Model)}} \\
\midrule

I'm trying to \orth{organize} \morph{my} \lex{linen closet} \morph{better, but it's such} a mess right now. \morph{Do you have} any tips for keeping sheets neat and accessible? &

I'm trying to organize \morph{me} linen closet \morph{where it's, like,} such a mess right now \morph{more good}.
\morph{Thou} have any tips for keeping sheets neat and accessible? &

I'm trying to \orth{organise} \morph{me} \lex{hot press} better, but it's such a mess right now.
\morph{What tips ye have} for keeping sheets neat and accessible? \\

\bottomrule
\end{tabular}%
}}

\label{tab:dialect-transformations}

\end{table}

To address these gaps, we introduce \methodName, a comprehensive framework for generating multi-turn, dialect-parallel dialogs. We partner with native linguists to identify the key lexical (vocabulary; parking lot $\rightarrow$ car park), orthographic (spelling; color $\rightarrow$ colour), and morphosyntactic (grammar;  Sarah and I $\rightarrow$ Sarah and meself) differences between SAE and their native dialects. We also ask them to categorize these features as appropriate or inappropriate for AI-generated model responses. Annotators indicate that up to 90\% of the eWAVE features should not be replicated in model responses, depending on the dialect.

Using SoTA LLMs, we generate seed dialogs and apply the rule-guided transformations identified as appropriate by native linguists. After a series of post-transformation quality checks, \methodName creates a collection of dialect-accurate multi-turn conversations alternating between user requests and model responses. In total, we produce 3,680 dialogs parallel across 9 English dialects--American (SAE; US), Australian (AU), British (GB), Canadian (CA), Indian (IN), Irish (IE), Nigerian (NG), Philippine (PH), and Scottish (SC)--with four turn variants (1, 2, 4, and 8) and three different datasets, resulting in a total of approximately 400k generated dialogs.

Using this data, we construct DialectLLM-Bench, spanning 9 English dialect variants to benchmark LLM dialect classification and response generation capabilities. Although dialect identification is a fundamental prerequisite to understanding dialectal input and responding appropriately, little research has examined models' capacity to discern English dialects~\cite{dialect_survey}. We present LLMs with our generated data and ask how well the model can a) identify the dialect and b) select the dialect-appropriate model response. Probing 17 LLMs across diverse families and sizes, we observe that even SoTA LLMs achieve only around 70\% accuracy on average on both tasks. These deficiencies are even more acute for certain dialects, as all tested models fail to reach 40\% average accuracy for CA. Examining failure patterns, we see a strong bias towards US (for CA \& PH) or GB (for IE, SC, \& IN) dialects. Collectively, this work suggests that dialect misidentification forms as an early failure point that can propagate to many observed non-SAE defects in tasks such as question-answering or reasoning. 

We further demonstrate that post-training on \methodName data substantially improves dialectal capabilities in LLMs. Notably, fine-tuning for dialectal response generation induces strong gains in dialect identification, suggesting that models acquire internal representations of systematic variation. The gains from training also extend to naturally occurring text. On a filtered subset of the International Corpus of English \cite{greenbaum1996international}, Qwen3-8B fine-tuned on DialectLLM data achieves 58.3\% average accuracy on dialect classification, compared to 38.9\% for the base model. This out-of-sample improvement suggests that model learns dialectal cues that transfer to real-world text.

In summary, (1) we propose \textbf{\methodName}, a novel dialect-aware data generation pipeline along lexical, orthographic, and morphosyntactic dimensions, using LLM-based transformations with native linguist annotations. (2) We explicitly delineate between morphosyntactic features common in users' dialects from those the model should use when generating text. (3) We construct a large-scale parallel multi-dialectal conversational dataset spanning 9 English dialects, enabling future research on dialect-aware conversational AI. (4) We demonstrate the utility of \methodName data for both benchmarking and post-training: \textbf{DialectLLM-Bench} shows the current LLMs' inability on dialect identification and generation, while post-training on \methodName data improves dialectal capabilities.

\section{Related Work}
\label{sec:related_work}

Dialect boundaries are inherently fluid, but in this paper we define dialects at the country level. Although regional and sub-regional differences exist within countries, this high-level categorization captures meaningful variation across the world. Given a dearth of high-quality dialect-specific data, particularly for lower-resource English dialects, substantial effort has been made to transform SAE text into target dialects. Hiring native speakers to manually transform text into their local dialect is effective but financially prohibitive to scale~\cite{ziems-etal-2022-value,lin-etal-2025-assessing}. As a result, most techniques transform baseline text by (a) using LLMs' parametric knowledge, (b) applying rule guided transformations (RGTs), or a combination of the two.

Parametric transformations are the simplest and commonly used~\cite{gupta-etal-2025-endive,finch-etal-2025-finding}, but risk reproducing the very biases that these data aim to correct~\cite{fleisig-etal-2024-linguistic}. Instead, most work applies RGTs with deterministic syntax parsing rules or an LLM. While some work targets lexical differences~\cite{zhou2025dialectgen}, most research exclusively focuses on the morphosyntactic features of a dialect, using eWAVE, a database of 235 morphosyntactic features used across 77 English varieties, compiled by professional linguists from 175 peer-reviewed sources~\cite{ewave}, as a source of ground truth rules \cite{ziems-etal-2022-value,multivalue,  xiao-etal-2023-tasklora,lee2025transenv, srirag-etal-2025-evaluating}. Its appeal is clear: comprehensive coverage, scientific grounding, and public availability.

Prior works \cite{ziems-etal-2022-value,multivalue} apply eWAVE using rule-based heuristics without LLMs, but these methods fail to capture the contextual depth of real-world dialect usages~\cite{lin-etal-2025-assessing} and often produce exaggerated feature applications (see App.~\ref{app:examples} for examples). \citet{gupta-etal-2025-endive} prompt GPT-4o with three eWAVE few-shot examples to transform SAE text to different dialects. \citet{lee2025transenv} provide GPT-4o-mini with eWAVE rules and LLM-generated guidelines alongside SAE text, using eWAVE's attestation ratings to assist with transformations. \methodName adopts a hybrid approach by using LLMs to apply native linguist validated lexical, orthographic, and morphosyntactic transformations, while uniquely distinguishing between features for model comprehension versus generation.

\section{\methodName}
\label{sec:data_generation}

We propose \methodName, a novel data generation framework to create a parallel conversational dataset across nine dialects. To properly integrate parametric knowledge from LLMs, rules from linguistic databases, and human-in-the-loop validation from native linguists, we decompose the transformation into five main components. This decomposition prevents LLMs from collapsing to SAE, which is often observed in modern LLMs~\cite{fleisig-etal-2024-linguistic}: (1) identify candidate transformations for SAE $\rightarrow$ target dialect for lexical, orthographic, and morphosyntactic dimensions, (2) validate these transformations with native linguists, (3) construct seed dialogs in SAE, (4) use LLMs to transform the seed dialogs via explicit transformation rules, and finally (5) apply quality controls to ensure accurate transformation. 

\begin{figure*}[t]

    \centering
    \includegraphics[width=0.99\linewidth]{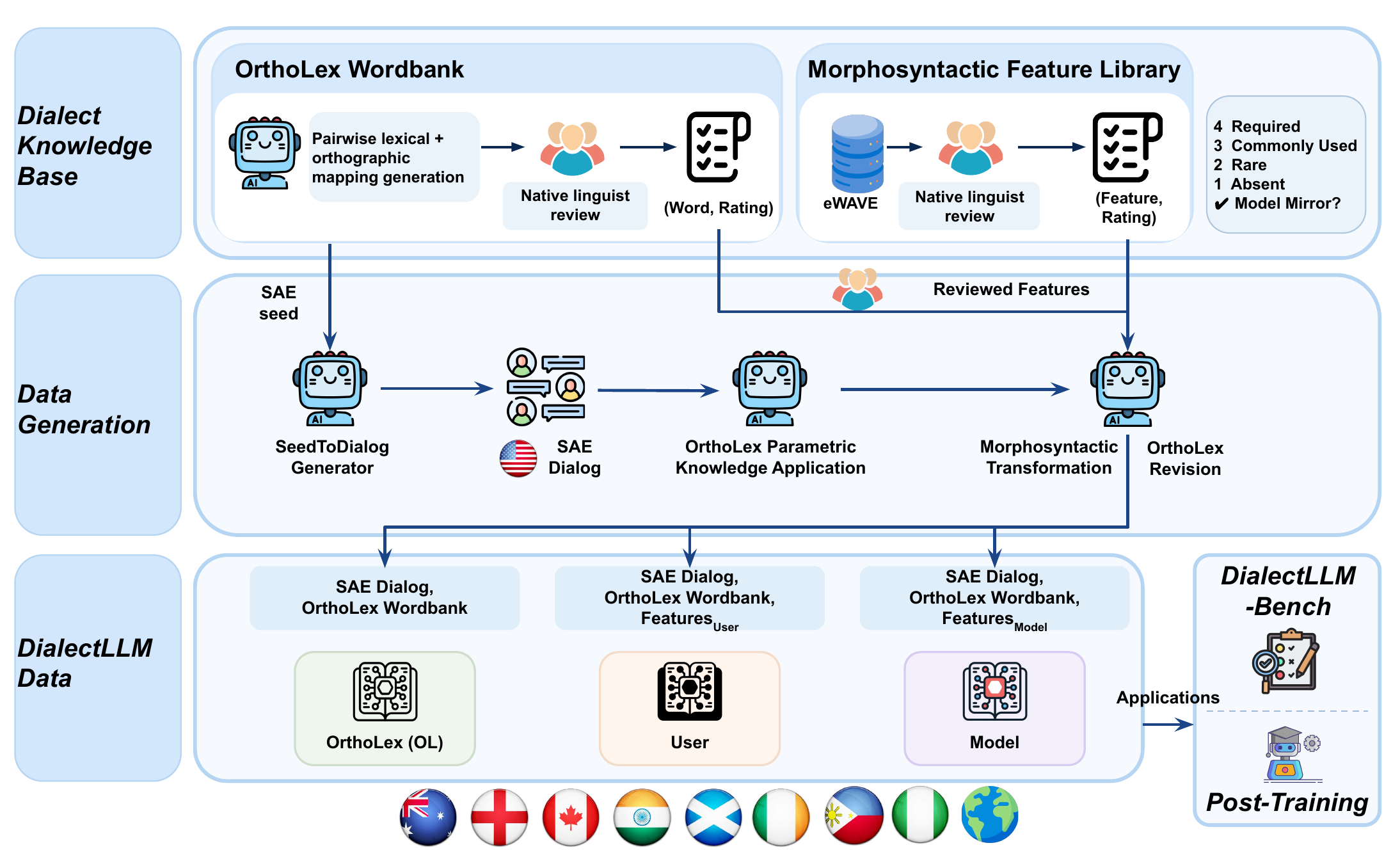}

    \caption{\textit{Overview of \methodName.} The framework combines ortho-lexical seed creation, multi-turn SAE dialog generation, and rule-guided transformation using linguist-curated lexical, orthographic, and morphosyntactic knowledge to produce parallel dialogs across 9 English dialects. The resulting dataset supports both benchmarking and post-training to enhance LLMs' dialectal capabilities.}
    \label{fig:initial_pipeline}

\end{figure*}

Our parallel construction preserves the underlying conversational content across dialects, while systematically varying linguistic features. This controlled design allows performance differences more directly to dialectal variations, rather than contextual cues such as location, currency, or cultural value. We visualize and formalize the entire generation process in Fig.~\ref{fig:initial_pipeline} and Algorithm~\ref{alg:mdial_pipeline}, respectively.

\subsection{Dialect Knowledge Base Construction}

\label{subsec:wordbank_construction}
The first stage of \methodName is to build a set of valid transformations from SAE to a target dialect. This consists of two sub-steps: (a) generating candidate transformations and (b) winnowing valid transformations from the candidate list to build the final set of transformations.

\noindent\textbf{Identifying Candidate Transformations.} Following prior research, we collect our candidate morphosyntactic transformations from eWAVE~\cite{ziems-etal-2022-value,multivalue,liu-etal-2023-dada,held-etal-2023-tada,xiao-etal-2023-tasklora,lee2025transenv}. Unfortunately, no similar comprehensive database exists to inform lexical and orthographic transformations for English dialects. To collect these additional dimensions, we first prompt GPT-5 and Gemini-3 with web search enabled to collect candidate mappings for each dialect. Then, with Claude-4-Sonnet as a verification filter, we validate each candidate as a genuine difference.

\noindent\textbf{Identifying Valid Transformations with Human Annotators.}
\label{subsec:human_annotation}
Critically, relying solely on LLM-based mappings may reinforce the very biases \methodName seeks to correct, hence we recruit native linguists to validate the authenticity of each feature transformation. Where possible, we recruit dialect-expert linguists; otherwise, we recruit three annotators from diverse backgrounds per dialect (e.g., our three Nigerian annotators belong to the Yoruba, Igbo, and Hausa ethnic groups; App.~\ref{subsec:annot_recruitment}).

Annotators rate each feature's prevalence using eWAVE's 1-4 scale, which we map to heuristic sampling probabilities: 4 = pervasive or obligatory (100\% frequency), 3 = neither pervasive nor extremely rare (60\%), 2 = exists but extremely rare (30\%), and 1 = attested absence (0\%).\footnote{We follow the heuristic probabilities of eWAVE~\cite{ewave,multivalue}.}

Maintaining an ordinal distinction is essential, as the SAE and dialect-specific features are not inherently mutually exclusive. For example, our British annotator gave ``call''$\rightarrow$``ring'' a 3, because both ``I'll call you'' and ``I'll ring you'' are natural in British English. Our scale captures this nuance.

For morphosyntax, our pilot experiments revealed that eWAVE does not reflect modern usage patterns and lacks some country-level dialects (e.g., no Canadian English or no Standard British but ``Southwest England'', ``East Anglian'', etc.). As a result, we have our linguists re-annotate each eWAVE feature for all target dialects. For dialects present in eWAVE, we exclude features marked as absent in the original eWAVE database to reduce workload.

Finally, we have linguists annotate whether the model should follow each feature when generating text. For instance, the discourse marker ``like'' (e.g., I am, \textit{like}, really tired.) was universally annotated as prevalent but not mandatory (3/4). At the same time, every annotator also indicated that models \textit{should not} adapt this feature. This novel annotation dimension uniquely distinguishes \methodName from previous approaches. 

\begin{figure}[!t]
    \centering
    \includegraphics[width=0.4\linewidth]{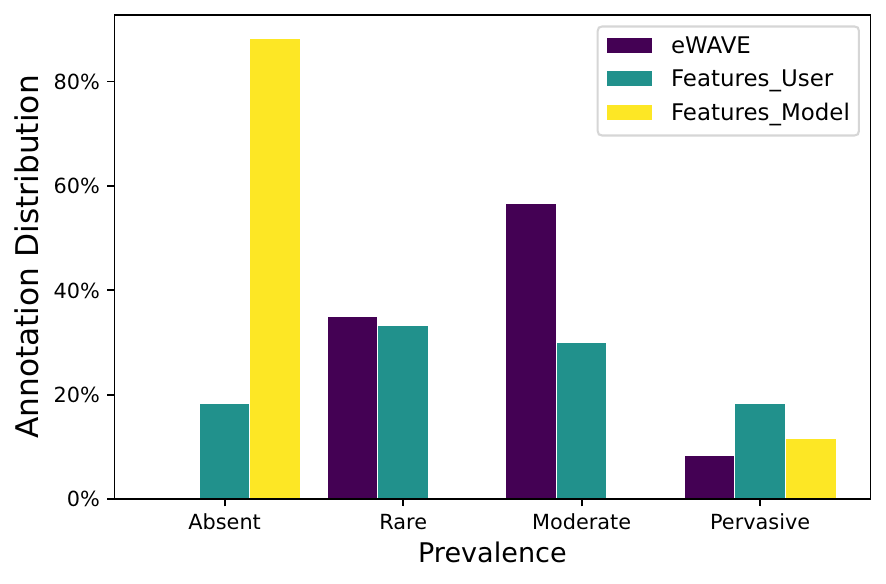}
    \hspace{0.02\linewidth}
    \includegraphics[width=0.4\linewidth]{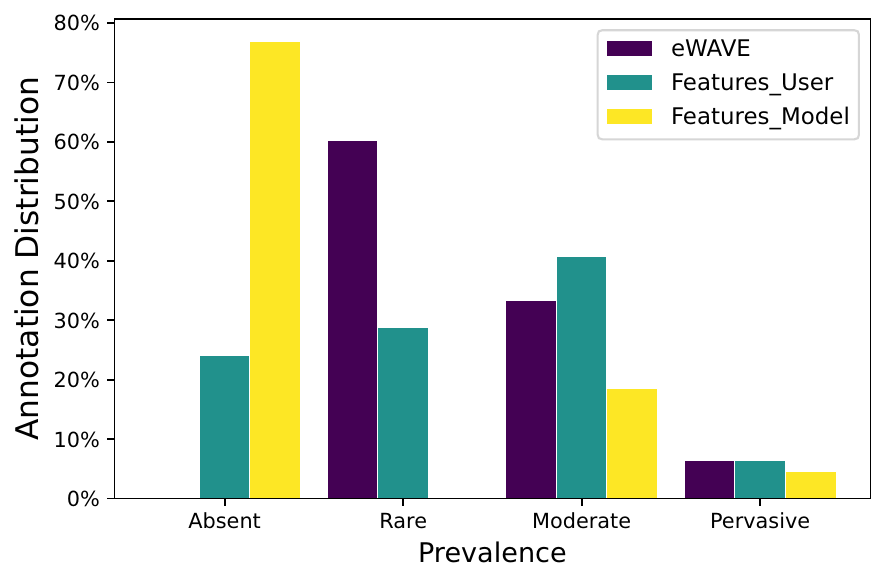}

    \caption{Prevalence rating distributions for AU (left) and PH (right) morphosyntactic features, comparing original eWAVE ratings with our newly annotated user- and model-appropriate ratings.}
 
    \label{fig:au_ph_annot_comp}
\end{figure}

Empirically, we see massive differences between eWAVE, user-appropriate ($\textit{Features}_{\text{User}}$), and model-appropriate ($\textit{Features}_{\text{Model}}$) morphosyntactic transformations; with per-feature alignment rate (agreement between eWAVE and annotator ratings) of only 34\% and 16\%, respectively. Moreover, our annotators indicate that models \textit{should not produce} up to 90\% of eWAVE features (Fig. \ref{fig:au_ph_annot_comp}). We also validate this distinction at the response level in Sec.~\ref{subsec:data_quality_analysis}.
These results clearly highlight a large gap in prior work, which has taken eWAVE annotations at face-value. Our work provides the first effort to disentangle the morphosyntactic features of a spoken dialect from how LLMs should generate text.

\subsection{Initial Dialog Generation}
\label{subsec:seed_generation}

We generate parallel dialect-accurate conversations by transforming source SAE text into other dialects with \methodName. To ensure the SAE dialects possess features which differ from non-SAE dialects, we synthetically generate baseline SAE dialogs using the seed words identified in Sec. \ref{subsec:human_annotation} as linguistically differentiating terms. We prompt SOTA LLMs to simulate a conversation between a user and an LLM agent based on the seed word. In \textbf{natural} conversations, the seed appears organically in the user's request (e.g., \textit{``Will I need an umbrella today?'')}, whereas in \textbf{indirect} conversations the user's initial request does not contain the seed and coaxes the model to generate it instead (e.g., ``\textit{What should I use to protect myself from the rain?'')}.

\subsection{Dialog Dialect Transformation}
\label{subsec:dialog_transform}

\methodName creates three datasets under different transformation settings: 
DialectLLM-OrthoLex (DialectLLM-OL), DialectLLM-User, and 
DialectLLM-Model. For each baseline dialog and dialect, \textit{d}, we sample the set of valid transformations using weights assigned to the 1-4 scale. For example, ``call''$\rightarrow$``ring'' was given a 3 in British English, corresponding to 60\% frequency. For each dialog, there is a 60\% chance this rule would be sampled. Note that most dialogs would not have this transformation applied either way, given the odds any single dialog contains ``call'' is quite low.

We found LLMs effectively apply lexical and orthographic rule-based transformations, both consistently applying the transformation rules while preserving semantic meaning and standard SAE grammar. We first transform the SAE dialog into the target dialect without explicit guidelines to fully utilize the LLM's parametric knowledge, then apply the revision process described in Sec.~\ref{subsec:qc_revision}. We conduct these steps to create DialectLLM-OL.

In contrast, we found that even SoTA LLMs demonstrate a strong tendency to default to grammatical patterns that are accepted in SAE even when explicitly prompted otherwise. Thus, rather than providing all morphosyntactic rules at once, we follow prior work and sequentially apply each morphosyntactic rule one-by-one~\cite{multivalue,lee2025transenv}. This stepwise procedure helps ensure that each transformation is faithfully applied. Importantly, we have two separate morphosyntactic transformation guidelines, one for users ($\textit{features}_{\text{User}}$) and a second for model ($\textit{features}_{\text{Model}}$) generation. Thus, we conduct this process twice to create DialectLLM-User and DialectLLM-Model for their respective guidelines.

\subsection{Quality Control \& Revision}

\label{subsec:qc_revision}

%Despite LLMs' general capabilities of executing proper lexical and orthographic transformations, we observed numerous errors and exaggerations. 

To ensure high-fidelity transformation, we engaged in iterative refinement in response to various challenges, which we discuss below.

\noindent\textbf{eWAVE Guideline Regeneration.} After providing example dialogs for DialectLLM-User \& DialectLLM-Model to native linguists, many felt that the grammatical transformations were still too extreme or incorrectly applied. We traced these errors to the limitations of eWAVE. The database does not tailor specific features to individual dialects and as a result fails to provide adequately detailed transformation guidelines. For example, the eWave rule for ``second person plural pronoun other than you'' allows many different transformations (y'all, youse, all of you, you guys, etc.) while only one or two options may be legitimate. This deficit creates awkward or unnatural text: y'all is used where youse is correct or vice versa.

Plus, we observe that prior academic work~\cite{lee2025transenv} that attempts to generate transformation guidelines/examples with LLMs introduced errors, misdirecting models to produce non-natural or inauthentic dialectal outputs. For instance, for an eWAVE rule, ``Leveling of the difference between present perfect and simple past: present perfect for StE simple past”, the provided example is ``She visited the museum yesterday'' $\rightarrow$ ``She has visited the museum.'', where ``yesterday'' is unintentionally dropped. When an LLM is given this guide during the rule application step, it incorrectly learns to drop temporal expressions entirely rather than understanding that the rule concerns tense leveling. As a result, the model would erroneously drop any temporal expressions leading to unintentional misguidance (e.g., ``It rained on Monday.'' $\rightarrow$ ``It has rained.''). To correct these salient issues, we regenerate the guidelines and examples with iterative interactions and refinement processes with native linguists. This procedure is essential, considering that LLMs are very sensitive to the given examples and guidelines~\cite{chatterjee-etal-2024-posix}.

\noindent\textbf{Orthographic \& Lexical Refinement.}
Another critical issue is that our annotations are inexhaustive, meaning they do not cover all possible transformations---many valid transformations are not represented in the original Knowledge Base. We rely on LLMs to fill in these gaps parametrically, but LLMs occasionally have wrong knowledge about the lexical and orthographic differences between dialects, even in high-resource dialects. For instance, an LLM transforms \textit{``eggplant''} to \textit{``aubergine''} for AU dialect, probably due to its influence from GB, while \textit{``eggplant''} is accepted in AU. Similarly, LLMs erroneously transform \textit{``-ize''} conventions (e.g., organize) to \textit{``-ise''} conventions (e.g., organise) for CA. We suspect that this occurs since CA follows GB spelling conventions with an exception of using \textit{``-ize''}. Moreover, LLMs frequently apply GB spelling conventions for PH, despite PH following SAE orthography. We hypothesize that this error stems from lack of knowledge of low-resource dialects. We discuss our further revision steps to ensure accurate transformations in App. \ref{subsec:further_data_quality}.

\section{\benchmarkName}
\label{sec:dialectllm-bench}
As a downstream application of \methodName, we introduce \benchmarkName, a multi-dialectal benchmark to evaluate LLM capabilities on dialect identification and response completion tasks.

\subsection{Classification: Dialect Identification}

\label{subsec:exp_classification}
For LLMs to be effective in dialect-aware interactions, they must first correctly identify the user's dialect from the conversational context. Imagine that a shopping LLM agent is asked ``Please purchase red pants size medium.'' If the user is British, they are asking the agent to purchase underwear while an SAE model would add trousers instead. If the model knew a-priori that the user is British, such a failure would be much less likely. Put simply, dialect identification failures can cascade to more severe failures. Though impossible to glean from this one sentence example, users naturally reveal their dialect over multi-turn conversations through their vocabulary, spelling, and grammar choices. 

Can models discern the English dialect from natural language? Although an integral task, dedicated benchmarks for English dialect detection remain scarce~\cite{dialect_survey}. This limitation stems from the scarcity of accurate, dialect-parallel datasets--a gap we address with \methodName. Through \benchmarkName, we evaluate this capability by measuring identification accuracy across varying conversational turns, using \{1,2,4,8\}-turn dialogs from \methodName data. Models are given a dialog and prompted to choose the correct dialect from a set of options. We provide the prompt template in App.~\ref{app:prompt_identification}.

\subsection{Generation: Response Completion}
\label{subsec:exp_response_completion}

Just as dialect identification is a necessary prerequisite to understanding the user's request, dialect-appropriate generation is essential to meet the needs of non-SAE speakers~\cite{why_AI_dialect}. However, evaluating open-ended generation in this context is challenging. Traditional n-gram metrics like BLEU~\cite{bleu_score} are often poorly correlated with dialect appropriateness given the breadth of the valid output space; LLM -as-a-judge is unreliable, considering the lack of dialectal knowledge of LLMs (Details in App.~\ref{app:llm_generation}).

Hence, we develop a response completion task---given a conversation history and the user's current turn request, select the correct model response from N candidates, each representing a different dialect---and treat this task as a proxy for response generation capabilities.  Note that we design the test set so that only one answer is valid in this task using the multi-label assignment mentioned in \ref{subsec:multi_label_assignment}. For the prompt template, see App.~\ref{app:prompt_response_completion}.

\subsection{Multi-label Assignment}
\label{subsec:multi_label_assignment}

\methodName transforms an SAE dialect into a single target dialect. However, a conversation may be valid in multiple dialects, especially short ones where dialectal cues are limited. This creates a challenge for dialect identification and response completion tasks where multiple options may be valid choices. Addressing this challenge, we construct a $O\times L \times M \times D$ matrix (orthographic $\times$ lexical $\times$ morphosyntactic $\times$ dialect) for all valid transformations. When generating data, we track every transformation and validate each dialect against this set of transformations. We assign secondary labels to that dialog if \textit{every} transformation is valid in additional dialects. 

\section{Experiments}
\label{sec:results}

\subsection{Models and Dataset}

We use 500 dialogs per turn (1, 2, 4, 8) and dataset (DialectLLM-\{OL, User, Model\}) resulting in approximately 54K test samples for evaluation (500 dialogs$\times$ 9 dialects$\times$ 3 datasets$\times$ 4 turn variants). We evaluate seventeen models, including Claude-$\{$Haiku-4.5, Sonnet-4, Opus-4.1$\}$, Qwen3-$\{$0.6B, 1.7B, 4B, 14B, 8B, 32B, 235B-A22B$\}$~\cite{yang2025qwen3}, Gemma3-$\{$4B, 12B, 27B$\}$~\cite{team2025gemma}, GPT-OSS-$\{$20B, 120B$\}$~\cite{agarwal2025gpt}, and Deepseek-$\{$V3.1, R1$\}$~\cite{liu2024deepseek,Guo_2025}. We also conduct supervised fine-tuning on Qwen3-8B and Gemma3-12B to see whether post-training enhances the dialectal capabilities of LLMs.

%We distribute 500 and 2500 dialogs per turn per dataset (\textit{OrthoLex}, \textit{RBT\_{user}}, and \textit{RBT\_{model}}) for evaluation and training, respectively. As a result, our test set comprises of approximately 54,000 test samples (500 conversations $\times$ 9 dialects $\times$ 3 datasets $\times$ 4 turn lengths) and 365,000 training samples after filtration. We use the dialogs in the test set and use them for tasks in MDialBench. In this paper, all data are generated by Claude-4-Sonnet.

\subsection{\methodName Quality Analysis}
\label{subsec:data_quality_analysis}

\begin{table*}[t]
\centering
\caption{
Native-speaker preference evaluations.
(a) \methodName win rate over Trans-EnV~\cite{lee2025transenv}.
(b) Preference between DialectLLM-Model and DialectLLM-User for assistant (LLM) responses.
}
\label{tab:data_quality_preferences}

\begingroup
\small
\setlength{\tabcolsep}{3pt}
\renewcommand{\arraystretch}{1.08}

\begin{minipage}[t]{0.49\textwidth}
\centering
\textbf{(a)} \methodName \textbf{vs.} Trans-EnV\par\vspace{0.35em}
\begin{tabular}{@{}lrrrrrr@{}}
\toprule
Eval. target & AU & IN & IR & NG & PH & SC \\
\midrule
User utt.   & 0.99  & 0.99  & 0.97 & 0.98 & 1.00 & 0.99 \\
Asst. resp. & 1.00 & 1.00 & 0.99 & 0.96 & 1.00 & 0.98 \\
\bottomrule
\end{tabular}
\end{minipage}
\hfill
\begin{minipage}[t]{0.49\textwidth}
\centering
(b) DialectLLM-Model (\texttt{M}) vs. DialectLLM-User (\texttt{U}) \par\vspace{0.35em}
\begin{tabular}{@{}lrrrrrr@{}}
\toprule
Preferred & AU & IN & IR & NG & PH & SC \\
\midrule
\texttt{M} & 0.80 &  0.53 &  0.99 &  0.91 &  0.88 & 0.74 \\
\texttt{U}  & 0.01  & 0.06  & 0.01  & 0.08  & 0.05  & 0.04  \\
\bottomrule
\end{tabular}
\end{minipage}
\endgroup

\end{table*}

\noindent\textbf{\methodName vs Prior Work.} We evaluate \methodName's data quality against data generated by the most recent SoTA method, Trans-EnV~\cite{lee2025transenv}, which provides a pipeline for synthetically generating non-SAE dialects. We construct single-turn SAE user-assistant dialogs and transform them into six non-SAE dialects shared by both frameworks. For each dialect, we recruit three native speakers, independent of the native linguists in Sec. \ref{sec:data_generation}, and ask which transformed dialog better represents (1) how a user would communicate and (2) how a model should respond in their English dialect. Preferences are aggregated via the median and we report the percentage of cases in which \methodName is preferred over Trans-EnV. \methodName substantially outperforms prior work, being preferred in approximately 99\% of cases for both user and model utterances (Table~\ref{tab:data_quality_preferences}(a)). Annotators consistently note awkward or incorrect grammatical transformations from Trans-EnV and compliment \methodName for attending to lexical and orthographic differences. See App.~\ref{app:preference_annotations} for full details.

\noindent\textbf{Validating User--Model Split.} We repeat this annotation procedure by comparing DialectLLM-User (\texttt{U}) and DialectLLM-Model (\texttt{M}) responses, this time asking annotators to focus exclusively on which response they prefer as a model response. Across six dialects, annotators prefer \texttt{M} in 80.65\% of pairs on average, while \texttt{U} is preferred in only 4.23\% (Table~\ref{tab:data_quality_preferences} (b)). The speakers comment that \texttt{U} responses apply informal or exaggerated features (e.g., ``ye/youse'', filler word such as ``like'', zero-article usage), that users may produce but assistants should avoid, hence greatly prefer \texttt{M} responses. Demonstrating remarkable consistency, features flagged by  annotators in \texttt{U} responses as problematic were precisely the features identified by our native linguists as inappropriate for model-generated responses when building the transformation pipeline. See App.~\ref{app:user_model_human_detail} for full details.

\subsection{LLM Performance on \benchmarkName}
\label{subsec:main_results}

\begin{figure}[!t]
    \centering
    \includegraphics[width=.9\linewidth]{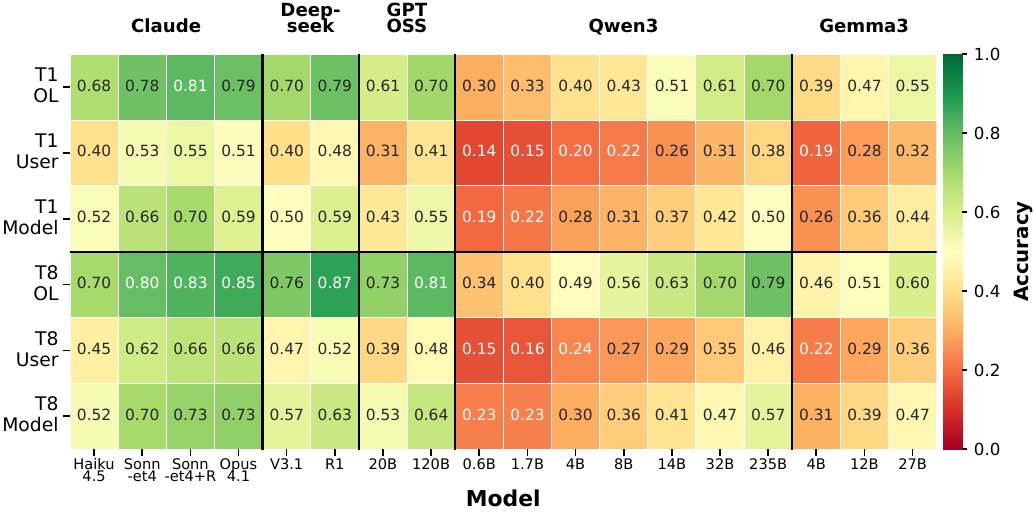}

    \caption{Model performance on \benchmarkName generation task for turn 1 and 8 dialogs. We observe an overall trend of positive correlation between model size \& accuracy and that models struggle when morphosyntactic features are added. ``+R'' indicates the model is equipped with reasoning. OL, User, and Model denote DialectLLM-OL, -User, and -Model dataset, respectively. Results across all turns for classification and generation are shown in Fig.~\ref{fig:classification} and Fig.~\ref{fig:response_completion}, respectively.}

    \label{fig:classification_part}
\end{figure}

Turning our attention now to \benchmarkName performance, we find that even the frontier models such as Claude-Opus-4.1 or Claude-Sonnet-4 (with reasoning) struggle with dialect identification and generation tasks (Fig.~\ref{fig:classification_part}). This finding illuminates prior work on model failures when engaging with non-SAE dialects ~\cite{ziems-etal-2022-value, fleisig-etal-2024-linguistic, lee2025transenv}, by highlighting a clear failure mechanism. If a model cannot infer whether a user is speaking Canadian, Nigerian, or Philippine English, it cannot reliably understand the user's request or adapt its response. Across the board, models perform worse on DialectLLM-User where morphosyntactic transformations are most extreme. Interestingly for classification (Fig. \ref{fig:classification}), models perform better on DialectLLM-Model than DialectLLM-OL, indicating that constrained morphosyntactic differences add information. Paradoxically, performance for small models \textit{decreases} for the classification task as context (turn count) increases. To assess the statistical reliability of the results, we compute 10K bootstrap uncertainty estimates, where standard errors across 17 models are extremely low (mean: .0070, range: .0058-.0076). For instance, Claude-Opus-4.1 and Qwen3-8B have 95\% CIs of [.69, .72] and [.31, .34], supporting stability of our estimates. We explore the relationships between turn count, model size, and performance below and conduct ablations to evaluate whether test-time compute for reasoning improves performance in App.~\ref{subsec:reasoning}.

\subsection{Scaling Effects: Model Size vs Performance}
\label{subsec:model_size_performance}

We observe a clear positive linear correlation between model size and performance. We use estimated sizes for Claude-family models, which lack public parameter disclosures (Haiku: 20B; Sonnet: 175B; Opus: 2T)\footnote{\url{https://claude.ai/public/artifacts/0ecdfb83-807b-4481-8456-8605d48a356c}}. Fig.~\ref{fig:model_size_performance} shows that the performance of all models scale linearly with model size until  plateauing at the largest scales. Though intuitive, this finding is noteworthy as LLMs develop dialect classification and generation capabilities despite being trained on unlabeled text without explicit dialectal metadata.

\subsection{Context vs Performance}
\label{subsec:turn_performance}

\begin{figure*}[t]
    \centering
    \begin{subfigure}[t]{0.45\linewidth}
        \centering
        \includegraphics[width=\linewidth]{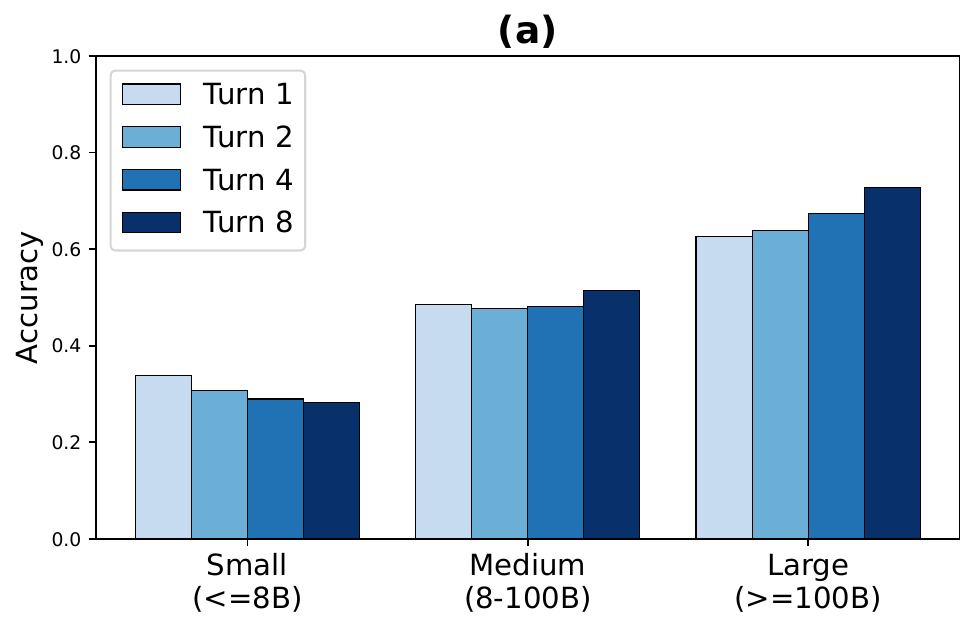}
    \end{subfigure}
    \begin{subfigure}[t]{0.45\linewidth}
        \centering
        \includegraphics[width=\linewidth]{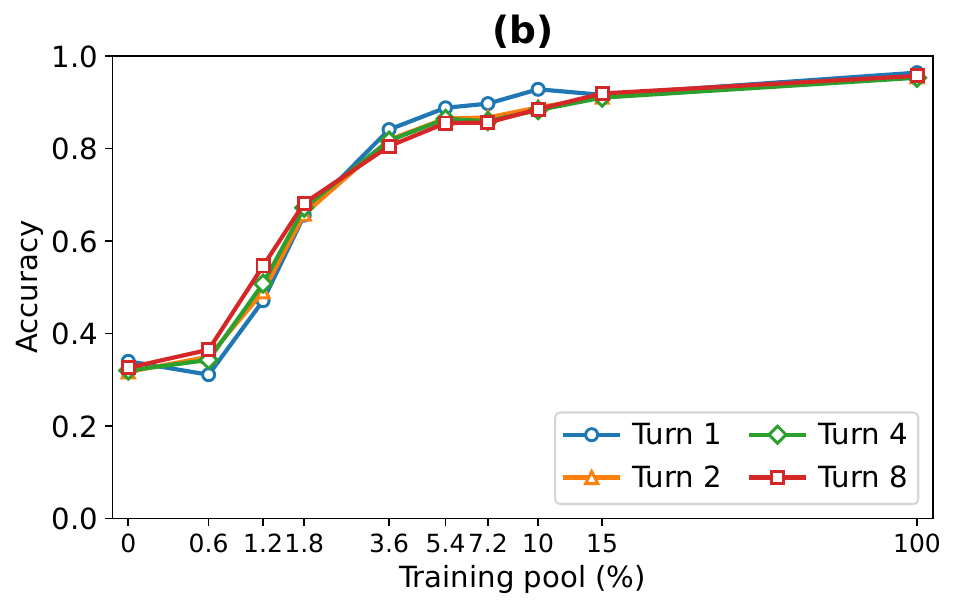} 
    \end{subfigure}

    \caption{
    (a) Model performance by scale and context length averaged over 
    all models and datasets for the classification task. (b) Qwen3-8B performance after fine-tuning on increasing fractions of the training pool for the generation task. The x-axis is displayed using $\log(1+x)$ spacing for readability.
    }

    \label{fig:context_length_performance}
\end{figure*}

We expect the performance of LLMs to improve with longer context (more turns), as more context provide clearer dialectal signals. Fig.~\ref{fig:context_length_performance}(a) shows a counter-intuitive pattern where smaller models perform better on short conversations. Investigating, we find that shorter conversations are often valid in multiple dialects (multi-label). Eight-turn dialogs, meanwhile, accumulate sufficient distinctive features and almost always has a single deterministic label.

Consequently, this ambiguity benefits models that make biased guesses. If a model defaults to high-resource dialects (e.g., SAE or British English), rather than genuinely analyzing dialectal features, it will score higher on ambiguous short conversations, where these dialects are often valid. All models smaller than 8B underperform the \textit{GB-biased guess} (always selecting GB as the answer; App. \ref{app:more_classification_generation}) for DialectLLM-User. Qwen3 models, in particular show large deficits on this task, with the 14B and 32B versions underperforming this baseline by 5\% accuracy.

We explore these failure patterns as confusion matrices of predicted and actual labels for each model (App.~\ref{app:turn_perf}). Smaller models demonstrate a consistent bias towards high-resource dialects, regardless of the true dialect, and their success on short conversations is an artifact of label ambiguity, not genuine capability. This bias persists in even the largest models and is particularly pronounced for Canadian English. When averaging performance across turns, Claude-Opus-4.1, for example, correctly classifies Canadian English just 33\% of the time, mislabeling 31\% of dialogs as US . This pattern holds for other large models---Deepseek-R1 (37\% correct; 27\% US), GPT-OSS-120B (31\% correct, 27\% US)---and is notably poor for Qwen3-235B (15\% correct, 69\% US) (Fig. \ref{fig:all_models_confusion_1} and \ref{fig:all_models_confusion_2}).

\subsection{Post Training with DialectLLM Data}
\label{subsec:post-training}

Can the dialogs from \methodName be used to improve LLM's dialectal capabilities? We conduct supervised fine-tuning (SFT) on Qwen3-8B and Gemma3-12B for both tasks, using 2,500 dialogs per turn and dataset, resulting in approximately 365K training samples after filtration. These training samples follow the same task formats as DialectLLM-Bench for classification and generation. We evaluate the fine-tuned models on \benchmarkName. To minimize data leakage, we construct seed-disjoint splits, ensuring that no seed word appears in both the training data and the test sets.

Fine-tuning substantially improves model performance (App.~\ref{app:post-training}), with turn 1 accuracy boosting from 30\% to approximately 90\%, surpassing frontier models. To verify that these gains do not simply reflect data memorization, we train Qwen3-8B on progressively smaller subsets of the training pool. Performance improves gradually as more training instances are added, where the turn 1 accuracy for the response completion rises from 47\% with 1.2\% of the training pool to 93\% with 10\% (Fig.~\ref{fig:context_length_performance}(b)). This trend suggests that \methodName data can help LLMs to learn comprehensive dialectal patterns.

\subsection{Cross-Task Transfer}
\label{subsec:cross-task}

Does fine-tuning on classification improve generation or vice versa? To observe potential transfer learning, we post-train models on each task separately and report the results. Models trained for response completion (generation) demonstrate improved classification performance (Table~\ref{tab:app_cross_identification}). For example, Qwen3-8B-RC-0.15 performance on turn 1 classification improves by 18.8, 16.0, and 19.5 percentage points (pp) for DialectLLM-User, -Model, and -OL, respectively.\footnote{We mainly report the numbers for models trained on 15\% of the training pool to avoid overfitting. Full results in App.~\ref{app:cross-task}.} In contrast, we see a 10.6 pp decrease on average in response completion accuracy after classification SFT (Table~\ref{tab:app_cross_rc}).

%for DialectLLM-User, 8.2 pp for DialectLLM-Model, and 17.0 pp for DialectLLM-OL

These findings suggest dialect identification capabilities naturally emerge as models learns the distinct features of different dialects when generating responses (Table ~\ref{tab:app_cross_identification}). Response completion requires fine-grained comparison between the candidates, building transferable discriminative capabilities. Classification, in contrast, involves assigning a single label to an entire dialog, a coarser task that may incur shortcut learning (e.g., detecting surface-level lexical cues) without developing a nuanced comparative reasoning required for response completion.

\subsection{Generalization Across Variants and Corpora}
\label{subsec:generalization_data_distribution}
We further examine whether the gains from post-training are robust to distribution shift by exploring cross-variant transfer (e.g., training the model on DialectLLM-OL and testing on -User) and dialect identification on naturally occurring text from the International Corpus of English (ICE)~\cite{greenbaum1996international}.

\noindent\textbf{Cross-variant Transfer.} As a controlled proxy for distribution shift, we train Qwen3-8B on each DialectLLM variant (-OL, -User, and -Model) and test on held out test sets of each variant. Every cross-variant result exceeds the base model with average gains of 34.3 and 41.8 pp in dialect identification and response completion, respectively. These results indicate transfer across dialect realizations rather than fit to one dataset variant (details on App. \ref{app:cross-variant}).

\noindent\textbf{Evaluation on Naturally Occurring Text.} We collect text from ICE, a corpus of written and spoken English, segmented by nationality. We test whether our post-trained model, trained on 9 dialects, outperforms the baseline on identifying the six overlapping dialects in the dataset (AU, NG, IN, IE, PH, US). We filter out text with clear location/cultural signals (e.g., Sydney, Naira [NG currency]), insufficient context, or no linguistic dialect markers. Following our benchmark design, we present the model with 9 multiple choice options and task it to choose the correct dialect. Our post-trained model improves upon the base model by 19.4 percentage-points in accuracy (38.9\% to 58.3\%) across the six dialects, providing both provide out-of-domain validation and highlight the value of our resulting dataset from DialectLLM. Experimental details and additional analyses are detailed in App.~\ref{app:ice_data}.

\section{Conclusion}
\label{sec:conclusion}

We introduce \methodName, a framework for generating parallel multi-dialectal English dialogs. The resulting data support a range of applications, including dialectal evaluation through \benchmarkName and LLM post-training. Our primary theoretical contribution is to compartmentalize the dialect features present in user utterances, which models should understand, and those the model should produce in textual responses. We validate this bifurcation with expert native linguists, many of whom specialize in dialect-appropriate model responses, and find that as many as 90\% of features in the eWAVE database \textit{should not} be reproduced by models. The absence of this distinction in prior work risks perpetuating stereotypical and exaggerated dialect representations. Whereas prior work targets a single dialect dimension such as vocabulary \cite{zhou2025dialectgen} or grammar \cite{ziems-etal-2022-value, xiao-etal-2023-tasklora, srirag-etal-2025-evaluating}, to our knowledge \methodName is the first generator to encompass all three dimensions of written dialects: lexical, orthographic, and morphosyntactic dimensions. The result is the most authentic and natural dialect representations to date, with annotators preferring \methodName over most recent prior work~\citep{lee2025transenv} around 99\% of the time. Using these data, we construct \benchmarkName to evaluate dialect identification and response completion. Surprisingly, even frontier LLMs struggle with both tasks, achieving around 70\% accuracy across different turns and exhibiting strong bias toward high-resource dialects (US and GB). Finally, we show that \methodName data also serve as a scalable post-training resource, boosting dialectal-task performance with gains that extend to naturally occurring text.

\noindent \textbf{Limitations \& Future Work.} \methodName focuses on transcribable, linguistic, and country-level variations across English dialects, aiming to preserve the semantic content of dialogs. We leave natural extensions, such as speech/audio-linked variation including accents and phonetics, finer-grained within-country variations, non-English language dialects, integration of cultural context to future work (more details on App. \ref{app:future_work}).

\begin{ack}
We thank Keira Daley, Shammie Debnath, Sean Mac Aodha, Aditi Garde, and Diana Cosma, whose annotations and iterative feedback shaped DialectLLM’s dialect knowledge base and transformation guidelines. We also thank Barbara Fero and Colleen Myers for their support and facilitation in recruiting linguistic annotators.
\end{ack}

\bibliography{main}
\bibliographystyle{unsrtnat}

%%%%%%%%%%%%%%%%%%%%%%%%%%%%%%%%%%%%%%%%%%%%%%%%%%%%%%%%%%%%%%%%%%%%%%%%%%%%%%%
%%%%%%%%%%%%%%%%%%%%%%%%%%%%%%%%%%%%%%%%%%%%%%%%%%%%%%%%%%%%%%%%%%%%%%%%%%%%%%%
% APPENDIX
%%%%%%%%%%%%%%%%%%%%%%%%%%%%%%%%%%%%%%%%%%%%%%%%%%%%%%%%%%%%%%%%%%%%%%%%%%%%%%%
%%%%%%%%%%%%%%%%%%%%%%%%%%%%%%%%%%%%%%%%%%%%%%%%%%%%%%%%%%%%%%%%%%%%%%%%%%%%%%%
\newpage
\appendix

\onecolumn

\section{Broader Impact}
\label{app:Broader_Impact}
\methodName provides a foundation for systematic study of LLM dialect capabilities, highlighting both the scale of current deficits and a path toward improvement. As conversational AI becomes ubiquitous, ensuring equitable performance across the world's English dialects is not merely a technical challenge but an ethical imperative. The over one billion non-SAE English speakers deserve AI systems that understand and respect their linguistic identities. \methodName offers an important component for any future research seeking to generate, benchmark, or train LLMs to support the multitude of English dialects.

\section{More Future Work}
\label{app:future_work}

Beyond country-level distinctions, dialect use varies across regions, communities, and individuals within a country. Combining DialectLLM's group-level  (country-level) personalization~\cite{zhang2025proper} with existing LLM personalization techniques~\cite{zhao2026llm} could help capture this finer-grained variation by accounting for both users' linguistic practices and their preferences for assistant responses. Such integration could support a broader range of users across diverse applications~\cite{zhang2024personalization,elwahsh2025linguistic,oh2026classroom}. A promising next step is to incorporate user feedback into the selection and application of dialect features according to individual preferences and interaction contexts.

Moreover, DialectLLM deliberately focuses on the linguistic, as opposed to cultural/semantic, dimensions of a dialect. We observe in our pilot experiments that cultural dimensions such as idioms rarely follow clean one-to-one transformation rules and adaptation often require changing culturally situated referents or the communicative scenario necessitating careful contextual expert judgment. We observed LLMs would introduce exaggerated stereotypes or US-centric assumptions, making the adapted text less appropriate for the target context.

As a result, DialectLLM conducts linguistic adaptation while preserving the semantic content of the source dialog through a set of probabilistic rule-based transformations. Rather than relying upon LLM’s parametric knowledge to fill in the gaps, we provide clear deterministic rule-based mappings to eliminate, to the extent possible, subjective culturally-biased transformations. However, the persistence of American cultural schemas embedded in the text is a genuine limitation.

In these domains, LLMs, which are heavily influenced by vast corpora from high-resource regions, demonstrate bias toward US-centric values and conventions, somewhat similar to the phenomenon where LLMs were biased towards high-resource dialects. Current efforts to mitigate these problems, focus on data generation frameworks~\cite{shi2024culturebank,li2024culturepark} and specialized post-training or prompting techniques~\cite{li2024culturellm,zhang2026mind} to foster more inclusive model behaviors. Integrating DialectLLM with research on cultural adaptation and evaluation~\cite{pawar2025survey} is therefore a promising direction for future work.

\clearpage
\section{Human Annotation Instructions }
\label{app:human_annotation_instruction}
\subsection{Lexical and Orthographic Annotations}
\begin{figure}[ht]
    \centering
    \begin{promptbox}{Instruction Template (Irish English)}
        \sffamily
        Task: Compare the original (US English) word/spelling with the transformed (Irish English) variant on a 1-4 scale:

        \vspace{0.8em}
        \begin{enumerate}[leftmargin=1.5em, nosep]
            \item \textbf{Feature is not used.} The transformed word/spelling is never correct in Irish English.
            \item \textbf{Feature exists but is rare.} The word/spelling might be understood but is outdated, region-specific, or influenced by other dialects. An LLM/voice agent should understand this term but not generate it.
            \item \textbf{Feature exists and is commonly used in colloquial speech.} The word/spelling is widespread in informal Irish English but may coexist with other variants. Both forms are acceptable.
            \item \textbf{Feature is required in this locale.} Using the US variant would sound unnatural or incorrect. The Irish variant is the standard and expected form.
        \end{enumerate}
        \vspace{1em}
        \textbf{Examples:}
        \vspace{0.5em}
        \scriptsize
        \begin{tabularx}{\linewidth}{|X|X|c|c|X|}
            \hline
            \textbf{US} & \textbf{IE} & \textbf{note} &\textbf{Rating} &\textbf{Explanation (Optional)}\\
            \hline
            Excellent, Great & Savage & lexical&& \\
            \hline
            Fruitcake & Barmbrack & lexical&& \\
            \hline
            -or (e.g., color, favor) & -our (e.g., colour, favour) & orthographic&& \\
            \hline
        \end{tabularx}
    \end{promptbox}

    \caption{Guidelines provided to Irish annotators for evaluating lexical and orthographic variations. Examples are truncated for brevity.}
\end{figure}

\clearpage
\subsection{Morphosyntactic Feature Annotations}

\begin{figure}[!ht]
    \centering
    \resizebox{0.92\linewidth}{!}{%
    \begin{minipage}{\linewidth}
    \begin{promptbox}{Instruction Template (Nigerian English)}
        \sffamily
        We are working with eWAVE, a database of English dialect grammars, to build models to generate Nigerian-sounding phrases. This is important for two reasons:
        
        \vspace{0.5em}
        1. \textbf{LLMs/conversational voice agents Understands non-US dialects:} Work with lexical variation found that LLMs/voice agents frequently misunderstood non-US terminology. When a user speaks with LLMs/voice agents, do they understand what the user is asking? Importantly, LLMs/voice agents should understand a variety of ways of speaking that are not necessarily “proper.” In the US, users might say “ain’t” in place of “is not,” or use “like” as a filler word when speaking. The model should understand that this word is purely filler and the user is not making a simile.
        
        \vspace{0.5em}
        2. \textbf{LLMs/voice agents respond with dialect-appropriate grammar:} Similarly, it is important that the models respond in a grammatically correct manner. 

        \vspace{0.8em}
        \textbf{Task} \\
        Using eWAVE’s list of grammar rules, we have created two versions of a sentence. The original sentence represents how the grammar rule in standard US English. The transformed sentence provides the sentence using this grammatical transformation. \textbf{Some example phrases are specific to certain regions. Please review the rules and explain any dialect-specific phrases that apply to them.} Please rate each rule by comparing the original and the transformed sentences on a 1-4 scale:

        \vspace{0.5em}
        \begin{enumerate}[leftmargin=1.5em, nosep]
            \item \textbf{Feature is not used.} The transformed sentence is never correct in this locale.
            \item \textbf{Feature exists but is rare.} It might be characteristic of a specific region or an antiquated way of speaking that is outdated. An LLM/voice agent should understand this phrasing. \\ \textit{US example: Double negatives: “He ain’t never said that”}
            \item \textbf{Feature exists and is commonly used in colloquial speech.} However, it is not required or even standard. It might be characteristic to a specific region of the country. \\ \textit{US example: Contractions: “ain’t” in place of “is not” or “cuz” in place of “because”}
            \item \textbf{Feature is required in this locale.} It would sound unnatural if the speaker used the original sentence instead of the transformed sentence. \\ \textit{US example: Possessive: The transformed sentence uses to “my object” when the original sentence uses “me object”}
        \end{enumerate}

        \vspace{0.8em}
        \textbf{[Tips for Model Mirror? Column]} \\
        For each feature listed, please evaluate LLM/voice agent's capabilities in the specified dialect and fill in the columns as follows:
        
        \vspace{0.3em}
        \textbf{Model Mirror? Column:} Write ``Y'' (Yes) or ``N'' (No) to indicate whether it would be better for the model to generate or respond with text or speech output including the feature for Nigerian English dialects.
        
        \vspace{0.5cm}
        
        \scriptsize
        \begin{tabularx}{\linewidth}{|X|X|X|X|c|c|}
            \hline
            \textbf{Transformation Rule} & \textbf{Original Sentence} & \textbf{Transformed Sentence} & \textbf{Description} & \textbf{Rating (1-4)} & \textbf{Model Mirror?} \\ \hline
            Me instead of I in coordinate subjects & Sarah and I are going to the market. & Sarah and me are going to the market. & This transformation rule replaces the subjective pronoun ``I'' with the objective pronoun ``me''... &  &  \\ \hline

        \end{tabularx}
    \end{promptbox}

    \caption{Guidelines provided to Nigerian annotators for evaluating morphosyntactic variations. Examples are truncated for brevity.}
\end{minipage}%
}
\end{figure}
\clearpage
\section{Details on Data Generation}

\subsection{Dialectal Transformation Features}
\begin{table}[h]
\centering
\caption{\textit{Dialectal transformation types in \methodName.} The table illustrates lexical, orthographic, and morphosyntactic variations that distinguish dialects.}
\vspace{0.1cm}

\resizebox{\linewidth}{!}{%
\begin{tabular}{|p{2.5cm}|p{8cm}|p{4cm}|}
    \hline
    \textbf{Transformation} & \textbf{Description} & \textbf{Examples} \\ \hline
    \textbf{Lexical} & Lexical transformations refer to differences in vocabulary or the specific words used to refer to the same concept. & - Soda / Pop \newline - Pants / Trousers \newline - Umbrella / Brolly \\ \hline
    \textbf{Orthographic} & Orthographic transformations involve variations in the spelling of words. The pronunciation and meaning are generally the same, but the written form differs. & - Color / Colour \newline - Theater / Theatre \newline - Realize / Realise \\ \hline
    \textbf{Morphosyntactic} & Morphosyntactic transformations are about differences in grammar, including how words are formed and how they are arranged in sentences. & - Y'all / Youse / You guys \newline - I like me shoes. \newline(I like my shoes in SAE) \\ \hline
\end{tabular}%
}
\end{table}

\begin{table}[h]
\centering
\caption{Examples of dialectal feature transformations across lexical, orthographic, and morphosyntactic features. Text in red indicates specific modifications of the corresponding feature.}
\small

\begin{tabular}{lp{10cm}}
\toprule
 & \textbf{Example Sentence} \\ \midrule
\textbf{Original (SAE)} & I realized the color of the theater poster was different from the sign, so we plan to get takeout at the gas station. Are you coming? (assuming the user is asking to two or more people) \\ \midrule
\textbf{+ Lexical} & I realized the color of the theater poster was different from the sign, so we plan to get \textbf{\textcolor{red}{takeaway}} at the \textbf{\textcolor{red}{servo}}. Are you coming? \\ \midrule
\textbf{+ Orthographic} & I \textbf{\textcolor{red}{realised}} the \textbf{\textcolor{red}{colour}} of the \textbf{\textcolor{red}{theatre}} poster was different from the sign, so we plan to get \textbf{takeaway} at the \textbf{servo}. Are you coming? \\ \midrule
\textbf{+ Morphosyntactic} & I \textbf{realised} the \textbf{colour} of the \textbf{theatre} poster was different \textbf{\textcolor{red}{to}} the sign, so we plan to get \textbf{takeaway} at the \textbf{servo}. Are \textbf{\textcolor{red}{youse}} coming? \\ \bottomrule
\end{tabular}
\label{tab:dialect_transform}
\end{table}

A visual depiction of the different transformations that occur in \methodName is shown in Table.~\ref{tab:dialect_transform}. Lexical transformations involve the substitution of vocabulary, such as replacing ``takeout'' to ``takeaway'' and ``gas station'' to ``servo''. Orthographic transformations modify spelling conventions, such as converting ``color'' to ``colour'' and ``realized'' to ``realised''. Finally, morphosyntactic transformations address grammatical variations, including prepositional changes (e.g., ``different from'' to ``different to'') and the adoption of dialectal pronouns like ``youse''.
\clearpage
\subsection{Examples of eWAVE Features}
\label{app:example_ewave_feature}

\newcolumntype{s}{>{\hsize=.6\hsize}X} % "Small" X column
\newcolumntype{b}{>{\hsize=1.8\hsize}X} % "Big" X column (Description)

\newcolumntype{L}{>{\hsize=2\hsize\raggedright\arraybackslash}X} 
% Define a standard width column for the others
\newcolumntype{S}{>{\hsize=0.5\hsize\raggedright\arraybackslash}X}
\begin{table}[h]
\centering
\caption{Exemplar eWAVE features. The corresponding examples and guidelines generated in \methodName with LLMs along with iterative refinement partnering with linguists.}
\begin{tabularx}{\textwidth}{@{} S S S L @{}} % 3 small columns, 1 big description column
\toprule
\textbf{Rule} & \textbf{Original} & \textbf{Transformed} & \textbf{Description} \\ \midrule

She/her used for inanimate referents & The car is beautiful, and it runs perfectly. & The car is beautiful, and she runs perfectly. & This transformation rule involves replacing neuter pronouns (it/its) with feminine pronouns (she/her/hers) when referring to inanimate objects. Instead of using the standard "it" to refer to non-living things like vehicles, machines, or objects, the transformed text personifies these items by assigning them feminine gender through pronoun usage. This creates a more personalized or affectionate tone toward inanimate objects, similar to how ships are traditionally referred to as "she" in maritime contexts. \\ \addlinespace

Me instead of I in coordinate subjects & Sarah and I are going to the market. & Sarah and me are going to the market. & This transformation rule replaces the subjective pronoun "I" with the objective pronoun "me" in coordinate subject constructions. When "I" appears as part of a compound subject joined by "and" (such as "Sarah and I"), it is changed to "me" (becoming "Sarah and me"), even though the pronoun is functioning as the subject of the sentence. This represents a departure from standard English grammar, which requires the subjective case "I" for subjects, in favor of using the objective case "me" in these coordinate constructions. \\ \addlinespace
Like as a focussing device&He was very angry.&He was, like, very angry.&This transformation involves inserting the word "like" as a discourse marker or hedge, typically surrounded by commas, to focus attention on or emphasize the phrase that follows. The word "like" functions as a parenthetical interjection rather than in its traditional grammatical roles, serving to highlight or draw special attention to the subsequent description or statement. This usage is characteristic of informal spoken English and creates a slight pause that signals the importance or precision of what comes next.\\ \bottomrule
\end{tabularx}
\end{table}

\subsection{Examples of Different Datasets}
\label{app:examples}
Table~\ref{tab:compare_dataset} illustrates the distinct characteristics of our three transformation approaches. \methodName-OL captures lexical substitutions (parking lot $\rightarrow$ car park, gas station $\rightarrow$ service station) and orthographic aspects (kilometers $\rightarrow$ kilometres) consistent with Australian English. DialectLLM-User incorporates features including plural forms of you (youse) and discourse markers like ``like'', resulting in heavily marked text that reflects the full range of attested dialectal features, while human annotators flagged that these features should not be mirrored by LLMs. \methodName-Model takes the middle ground by applying a subset of morphosyntactic features based on human annotation, which highlights our framework's distinction between features that should be understood versus generated.

\begin{table}[htb]
\caption{Comparison of example transformation from prior works (Multi-Value~\cite{multivalue} and Trans-EnV~\cite{lee2025transenv}) and \methodName-OL, \methodName-User, \methodName-Model datasets. Bold text denotes dialectal modifications from SAE for \methodName data.}
\centering
\resizebox{\linewidth}{!}{%
\begin{tabular}{lp{13cm}}
\toprule

\textbf{User Input} & I'm planning a cross-country road trip next month and I'm trying to figure out the best way to handle meals along the way. I don't want to rely entirely on fast food chains, but I also don't want to pack everything myself. What are some good options for finding decent food when you're driving through small towns or rural areas? \\ \midrule \midrule

\textbf{SAE} & That sounds like an exciting adventure! For road trips through small towns and rural areas, roadside restaurants are often your best bet for authentic, home-cooked meals. These local gems usually offer hearty portions at reasonable prices and give you a real taste of regional cuisine. Look for places with full parking lots - that's usually a good sign that locals eat there regularly. Many of these establishments are family-owned diners or cafes that have been serving travelers and community members for decades. You can also use apps like Yelp or Google Maps to find highly-rated local spots along your route, or ask for recommendations at gas stations when you stop to fuel up. \\ \midrule
\textbf{Multi-Value} & That sound an exciting adventure! Road trips small towns and rural areas, roadside restaurants are often your most good bet authentic, home-cooked meals. These local gems usually offers hearty portions reasonable prices and give you a real taste regional cuisine. Look places full parking lots - that's usually a good sign that eats there regularly. Many these establishments are family-owned diners or cafes has been serving travelers and community members decades. You can also use apps Yelp or Google Maps to find highly-rated local spots your route, or ask recommendations gas stations when you stops to fuel.\\ \midrule
\textbf{Trans-EnV}& That is sounding like a, like, exciting adventure, but! For road trips through small towns and rural areas, roadside restaurants, which usual offer, like, hearty portions at reasonable prices and give you a real taste of regional cuisines, are often your, like, goodest bet for authentic, home-cooked meals, but. Look for places with full parking lots, and avoid places with empty lots - those mustn't be good since locals clear don't eat there regular. There's many family-owned diner or cafes that served travelers and community members for decades. You can also use apps like Yelp or Google Maps to find, like, high-rated local spots along your route, or ask for recommendations at gas stations when you stop to fuel up. \\ \midrule
\textbf{DialectLLM-OL} & That sounds like an exciting adventure! For road trips through small towns and rural areas, roadside restaurants are often your best bet for authentic, home-cooked meals. These local gems usually offer hearty portions at reasonable prices and give you a real taste of regional cuisine. Look for places with \textbf{chockers} \textbf{car parks} - that's usually a good sign that locals eat there regularly. Many of these establishments are family-owned diners or cafes that have been serving \textbf{travellers} and community members for decades. You can also use apps like Yelp or Google Maps to find highly-rated local spots along your route, or ask for recommendations at \textbf{service stations} when you stop to fuel up. \\ \midrule
\textbf{DialectLLM-User} &  That \textbf{there} sounds like an exciting adventure! For road trips through small towns and rural areas, roadside restaurants are often your most best bet for \textbf{, like,} authentic, home-cooked meals. These local gems here usually offer hearty portions at reasonable prices and give youse a real taste of regional cuisines. Look for places with \textbf{chockers car parks} - that's usually a good sign that locals eat there regularly. Many of these establishments \textbf{here} are\textbf{, like,} family-owned diners or cafes that have been serving \textbf{travellers} and community members for decades. \textbf{Youse} can also use apps like Yelp or Google Maps to find more highly-rated local spots along your route, or ask for recommendations at \textbf{service stations} when \textbf{youse} stop to fuel up. \\ \midrule
\textbf{DialectLLM-Model} & That \textbf{there} sounds like an exciting adventure! For road trips through small towns and rural areas, roadside restaurants are often your best bet for authentic, home-cooked meals. These local gems here usually offer hearty portions at reasonable prices and give you a real taste of regional cuisine. Look for places with \textbf{chockers car parks} - that's usually a good sign that locals eat there regularly. Many of these establishments \textbf{here} are family-owned diners or cafes that have been serving \textbf{travellers} and community members for decades. You can also use apps like Yelp or Google Maps to find highly-rated local spots along your route, or ask for recommendations at \textbf{service stations} when you stop to fuel up. \\ \bottomrule
\end{tabular}%
\label{tab:compare_dataset}
}
\end{table}

\clearpage
% Uses the existing preamble: \usepackage{algorithm} + \usepackage{algorithmic} (+ amsmath)
% Do NOT also load algpseudocode -- loading both causes the "0:" line prefixes and the trailing "=0".
\subsection{\methodName Algorithm}
\label{app:mdial_algo}
\begin{algorithm}[!ht]
   \caption{\methodName Pipeline}
   \label{alg:mdial_pipeline}
\footnotesize
\begin{algorithmic}
   \STATE {\bfseries Input:} Source dialect $S$ (SAE), non-SAE dialects $\mathcal{T}$, eWAVE rules $\mathcal{R}$
   \STATE {\bfseries Output:} Multi-dialectal dataset $\mathcal{D}_{\mathrm{final}}$
   \STATE {\bfseries Initialize:} $\mathcal{D}_{\mathrm{src}}, \mathcal{D}_{\mathrm{final}} \leftarrow \emptyset$;\;
       sampling probabilities $(p_1,p_2,p_3,p_4) \leftarrow (0,0.3,0.6,1)$

   \STATE \textit{// Step 1: OrthoLex Wordbank Construction}
   \FORALL{$t \in \mathcal{T}$}
       \STATE $\mathcal{C}_t \leftarrow \mathrm{LLM}_{\mathrm{gen}}(\mathrm{RetrieveDiff}(S,t))$
       \STATE $\mathcal{W}_t \leftarrow \{(w_s,w_t) \in \mathcal{C}_t : \mathrm{LLM}_{\mathrm{ver}}(w_s,w_t)\}$
   \ENDFOR

   \STATE \textit{// Step 2: SAE Seed Dialog Generation}
   \STATE $\mathcal{W}_{\mathrm{seed}} \leftarrow \{w_s : (w_s,\cdot) \in \bigcup_{t\in\mathcal{T}} \mathcal{W}_t\}$
   \FORALL{$(w_s,k,m) \in \mathcal{W}_{\mathrm{seed}} \times \{1,2,4,8\} \times \{\mathrm{Natural},\mathrm{Indirect}\}$}
       \STATE $\mathcal{D}_{\mathrm{src}} \leftarrow \mathcal{D}_{\mathrm{src}} \cup \{\mathrm{GenDialog}(w_s,k,m)\}$
   \ENDFOR

   \STATE \textit{// Steps 3--4: Guideline Refinement and Human Annotation}
   \FORALL{$t \in \{S\} \cup \mathcal{T}$}
       \STATE $\mathcal{G}_t \leftarrow \mathrm{RefineGuidelines}(\mathcal{R},t)$
       \STATE $\mathcal{A}_{\mathrm{Morph},t} \leftarrow \mathrm{HumanAnnotation}(\mathcal{G}_t)$ \hfill (prevalence 1--4, ModelMirror)
       \IF{$t \neq S$}
           \STATE $R_{h,t} \leftarrow \mathrm{HumanAnnotation}(\mathcal{W}_t)$
           \STATE $\widehat{\mathcal{W}}_t \leftarrow \mathrm{CompleteWordbank}(\{\mathcal{W}_u\}_{u\in\mathcal{T}},t,R_{h,t})$
       \ENDIF
   \ENDFOR
   \STATE For an OrthoLex mapping $c$ in dialect $t$: $q_t(c)=p_{R_{h,t}(c)}$ if a human rating exists; otherwise
   \STATE $q_t(c)=1$ if $R_{m,t}(c)=4$ and $0$ otherwise, where $R_{m,t}$ is an LLM rating few-shot prompted with $R_{h,t}$.

   \FORALL{$(d,t) \in \mathcal{D}_{\mathrm{src}} \times (\{S\} \cup \mathcal{T})$}
       \STATE \textit{// Step 5: Initial OrthoLex Transformation}
       \STATE $d_0 \leftarrow \begin{cases} d, & t=S,\\ \mathrm{ApplyOrthoLex}(d,t,\mathcal{W}_t), & t\neq S. \end{cases}$
       \FORALL{$v \in \{\mathrm{OL},\mathrm{User},\mathrm{Model}\}$}
           \STATE $d' \leftarrow d_0$
           \STATE \textit{// Step 6: Variant-Specific Morphosyntactic Transformation}
           \IF{$v \neq \mathrm{OL}$}
               \FORALL{applicable rules $r \in \mathcal{A}_{\mathrm{Morph},t}$}
                   \IF{$(v=\mathrm{User} \lor r.\mathrm{ModelMirror}) \land \mathrm{Random}() < p_{r.\mathrm{prevalence}}$}
                       \STATE $d' \leftarrow \mathrm{ApplyRule}(d',r)$
                   \ENDIF
               \ENDFOR
           \ENDIF
           \STATE \textit{// Step 7: OrthoLex Quality Control and Revision}
           \IF{$t \neq S$}
               \STATE $C \leftarrow \mathrm{ExtractOrthoLexChanges}(d,d')$
               \FORALL{$c \in C$}
                   \STATE Revert $c$ in $d'$ with probability $1-q_t(c)$
               \ENDFOR
               \FORALL{applicable mappings $c \in \widehat{\mathcal{W}}_t \setminus C$}
                   \STATE Apply $c$ to $d'$ with probability $q_t(c)$
               \ENDFOR
           \ENDIF
           \STATE $\mathcal{D}_{\mathrm{final}} \leftarrow \mathcal{D}_{\mathrm{final}} \cup \{(\mathrm{id}(d),t,v,d')\}$
       \ENDFOR
   \ENDFOR
\end{algorithmic}
\end{algorithm}

\subsection{Annotator Recruitment and Assignment.} 
\label{subsec:annot_recruitment}

%We put significant effort on linguist recruitment for quality and language/ethnicity-wise diversity. In more detail, we ensure the quality of annotations with several attention checks and recruit people from diverse backgrounds to capture intra-country dialect variation, for instance, speaking Cebuano, Ilocano, and Tagalog along with English for PH and ethnicity of Yoruba, Igbo, and Hausa for NG.

%Our annotation design reflects a tradeoff between breadth and depth. 

To construct our knowledge base of features, we required annotations of a large and technically detailed lists of lexical, orthographic, and morphosyntactic features, as well as iterative revisions of transformation guidelines and generated examples. We therefore prioritized annotators with relevant experience developing locale-appropriate language. We devoted significant effort on linguist and native speaker recruitment with several attention checks for quality, while seeking diversity among annotators' linguistic and ethnic backgrounds.

When available (Canadian [CA], Australian [AU], British [GB], Indian [IN], and SAE [US]), we recruited and worked with linguistic experts with professional experience developing dialect-appropriate language for LLMs. This was not always possible, particularly for lower-resource dialects. To ensure inclusive coverage, we worked with a professional annotation firm to recruit local annotators or linguists of the locale, who have relevant experience in annotation, with heterogeneous backgrounds to account for intra-country dialect variation. For Philippine and Nigerian English, for instance, we deliberately recruited participants from different ethnolinguistic backgrounds—Cebuano, Ilocano, and Tagalog backgrounds for Philippine English, and Yoruba, Igbo, and Hausa backgrounds for Nigerian English.

Specifically, for Irish (IE), Scottish (SC), Indian (IN), Philippine (PH), and Nigerian (NG) English dialects, 3 native speakers or linguists per locale were recruited, with majority voting employed to determine the final labels for each attribute. For Canadian (CA), Australian (AU), British (GB), and SAE (US), one native linguist, who specialize in dialect-appropriate LLM generation, per locale was assigned with \textit{iterative feedback cycles} to ensure annotation quality and consistency.

To mitigate ego-centric bias, we instructed annotators to evaluate general linguistic usage in their country, rather than their own personal usage. Their comments also highlighted regional variation, such as ``this feature is common in Dublin Irish English.'' Across the five dialects evaluated by three annotators each, all three annotators agreed on whether a feature was present in 82.1\% of cases overall, with agreement ranging from 76.0\% to 89.4\% across dialects.

\subsection{Visual Depiction of Wordbank Aggregation}
Continuing from Sec.~\ref{subsec:qc_revision}, we show a visual depiction of the aggregated wordbank matrix filled by LLMs and humans.

\begin{figure}[htbp]
    \centering
    \includegraphics[width=0.6\linewidth]{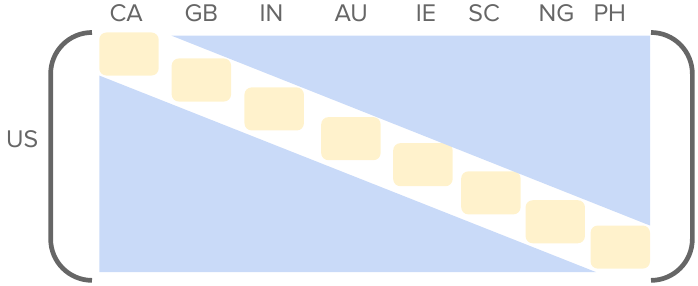}

    \caption{LLM-based wordbank matrix completion. Using human annotations as few-shot demonstrations (diagonal yellow), we prompt an LLM to fill out missing entries in the dialectal mapping matrix (blue regions).}

    \label{fig:wb_filling}
\end{figure}

\subsection{Annotation Rating Alignment and Distribution  (eWAVE vs \methodName)}
\begin{figure}[htbp]
    \centering
    \includegraphics[width=\linewidth]{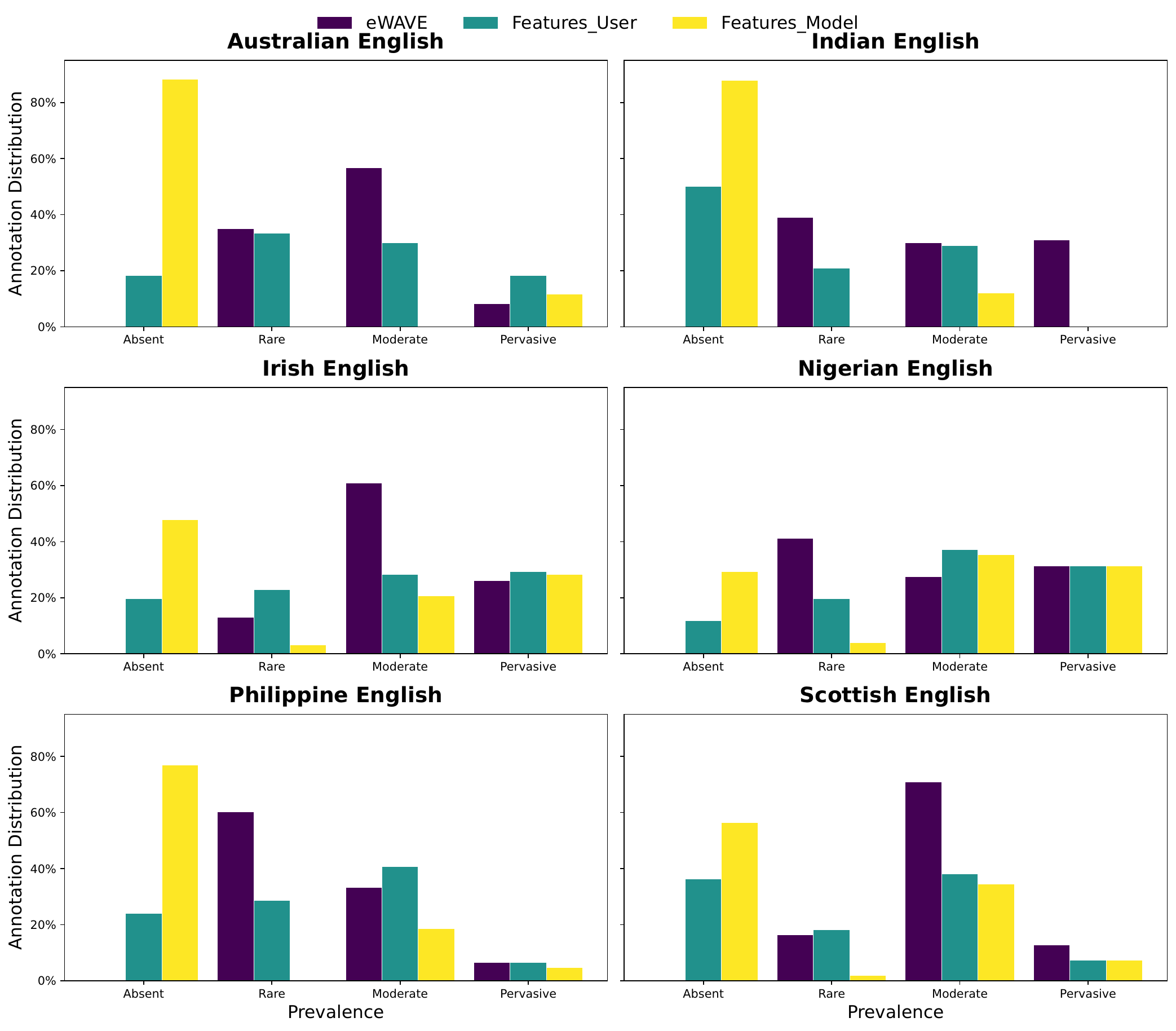}

    \caption{Annotation rating distribution of eWAVE, $Features_{User}$, and $Features_{Model}$ for dialects that exist in the eWAVE database. We observe clear discrepancies between the new ratings from native linguists and the original eWAVE annotations, with larger differences for $Features_{Model}$.}

    \label{fig:ewave_ours_comparison}
\end{figure}

Continuing from Sec.~\ref{subsec:human_annotation}, we report the feature-level alignment rate (Table~\ref{tab:ewave_align}) and the general distribution (Fig.~\ref{fig:ewave_ours_comparison}) of human ratings compared to the ratings present in the original eWAVE database. We observe a huge discrepancy in both alignment and distribution, where these differences become more pronounced for the $Features_{Model}$. Notably, the alignment gets as low as 18\% and 2\% alignment for $Features_{User}$ and $Features_{Model}$, respectively. This finding suggests that the mismatch can be the primary cause of the perceived unnaturalness in eWAVE-based generation.

\begin{table}[ht]
\centering
\vspace{0.5cm}
\caption{Alignment rates between eWAVE features and human annotations for $Feature_{User}$ and $Feature_{Model}$. Low alignment across all dialects underscore the inadequacy of using raw eWAVE ratings for natural dialect transformation.}
\label{tab:ewave_align}

\begin{tabular}{lcc}
\toprule
\textbf{Dialect} & \textbf{$Features_{User}$} (\%) & \textbf{$Features_{Model}$} (\%) \\
\midrule
Australian English & 41.67 & 1.67 \\
Indian English     & 18.00 & 2.00 \\
Irish English      & 40.22 & 27.17 \\
Nigerian English   & 41.18 & 31.37 \\
Philippine English & 33.33 & 7.41 \\
Scottish English   & 27.27 & 25.45 \\
\bottomrule
\end{tabular}

\end{table}

\clearpage

\subsection{Additional Data Quality Assurance}
\label{subsec:further_data_quality}
Continuing from Sec.~\ref{subsec:qc_revision}, first, we aggregate mappings present across all SAE $\leftrightarrow$ other dialects wordbanks. We provide an LLM human ratings from Sec.~\ref{subsec:human_annotation} as few-shot demonstrations and calibration reference and prompt to rate the missing part of the matrix (See Fig.~\ref{fig:wb_filling} for visual depiction and App.~\ref{app:qc_revision_prompt} for prompt templates). This matrix assists with comprehensive coverage of words. Second, to ensure that the initial transformations (fully by LLMs) are valid, an LLM, guided with human ratings as few-shot reference, is prompted to rate the transformations that have happened. Note that human ratings ($R_h$) are prioritized over LLM ratings ($R_m$) in all cases. For mappings that $R_h$ don't exist, we apply a conservative threshold, where only those with $R_m=4$ are accepted as valid transformations, while a rating of 1 to 3 results in reversion to the original SAE form.

If the initially executed transformation is rated 4 by humans or LLMs, the transformation is retained. If $R_h=3, 2$, or $1$, we revert the change probabilistically (40, 70, or 100\% respectively) and $R_m \le 3$, revert. For transformations that did not occur initially (false negatives with reference from wordbank, we apply them in a probabilistic manner ($R_h = 2,3,$ or $4$ $\rightarrow$ $30, 60,$ or $100\%$, respectively; $R_m=4$ $\rightarrow$ 100\%). As a result, we create 3 final datasets, \methodName-OL, \methodName-User, and \methodName-Model, where each dataset has different magnitude of morphosyntactic transformations. 

\subsection{Comparison to Prior Work: Human Preference Annotations}
\label{app:preference_annotations}

We compare our data generation method with Trans-EnV \cite{lee2025transenv}, a leading non-SAE dialect generator. We apply their method on the synthetically generated conversations described in Sec. \ref{sec:data_generation} using their open-source repository\footnote{https://github.com/jiyounglee-0523/TransEnV}. Off the shelf, we observed that their prompt struggled with the ``USER: ... ASSISTANT:...'' text format, so we modified the prompt slightly to include the following two lines:

\begin{quote}
- CRITICAL: Your output must contain EXACTLY ONE complete conversation with ONE ``USER:'' and ONE ``ASSISTANT:''. Do NOT include the original conversation in your output.\\
- CRITICAL: Your output must ALWAYS contain both USER: and ASSISTANT: sections, even if only one is transformed.
\end{quote}

With this change, we observe strong adherence to the model's prompt and a faithful application of the eWAVE transformation rules through Trans-EnV. 

We partner with a 3P data annotation service to recruit three annotators to compare 120 \methodName and Trans-EnV generated conversations. We present these two data transformations side-by-side and ask annotators to provide two preference annotations on 1 to 5 scale (one for the user utterance and a second for the model response):

\begin{quote}
\begin{enumerate}
    \item V1 is \textbf{greatly preferred} over V2
    \item V1 is \textbf{somewhat preferred} over V2
    \item No clear preference between V1 and V2
    \item V2 is \textbf{somewhat preferred} over V1
    \item V2 is \textbf{greatly preferred} over V1
\end{enumerate}
\end{quote}

Order was randomly shuffled to ensure no annotation bias. We provide the following additional annotation instructions:

\begin{quote}
\textbf{For the USER annotations:}
Which version is more representative of how users from your country speak in day-to-day conversation? We are trying to capture colloquial expressions (e.g., “y’all” instead of “you all”) and other features of the spoken dialect. These may include forms that are not technically “proper” but are commonly used and should be understood by the model.

\textbf{For the ASSISTANT annotations:}
Which model response is more appropriate for your country? While users may employ informal or nonstandard constructions when speaking (e.g., double negatives such as “ain’t got none”), the model generally should not. Use your best judgment to decide what a dialect-appropriate model response should sound like.

`Strong' indicates high confidence or a clear qualitative gap; `moderate' indicates a noticeable but smaller difference. Please try to avoid selecting (3) unless there is truly no clear difference. If both transformations are bad, prefer the option that is less bad. 
\end{quote}

We aggregate these preference annotations by taking the median value between the three annotators and present the percentage of time \methodName is preferred over Trans-EnV in Table \ref{tab:data_quality_preferences}(a). Table 
\ref{tab:preference_distribution} provides detailed results. Importantly, 99\% of the time \methodName is the preferred generator for both the User and the Model. Frequently, our annotators highlighted the awkwardness of the grammatical transformations of Trans-EnV and complimented the proper lexical and orthographic usages of \methodName, e.g., ``V2 (\methodName) is clear and natural, and uses the correct term “dickey” (common in Indian English). V1 (Trans-EnV) is grammatically broken and awkward, so it doesn’t sound like normal speech.''

Collectively, these findings emphasize the limitations of eWAVE as a source of dialect transformation guidelines. While both \methodName and Trans-EnV rely upon eWAVE as a ground truth source for morphosyntactic transformations, we worked with native linguists to heavily prune the set of valid transformations. As a result, \methodName imposes substantially fewer grammatical transformations, preserving the naturalness of the conversation. While any given transformation in eWAVE may be valid in the abstract, applying many rules simultaneously results in a grammatically fragmented and unnatural transformation.

\begin{table}[h]
\centering
\caption{\methodName vs Trans-EnV Preference Distribution (\%)}
\label{tab:preference_distribution}
\resizebox{\linewidth}{!}{%
\begin{tabular}{llcccccc}
\toprule
Role & Preference & AU & IN & IE & NG & PH & SC \\
\midrule
 & \methodName greatly preferred to Trans-EnV & 0.90 & 0.94 & 0.68 & 0.79 & 0.88 & 0.73 \\
 & \methodName somewhat preferred to Trans-EnV & 0.09 & 0.05 & 0.28 & 0.19 & 0.12 & 0.26 \\
User & \methodName approximately equal to Trans-EnV & 0.00 & 0.00 & 0.02 & 0.01 & 0.00 & 0.00 \\
 & Trans-EnV somewhat preferred to \methodName & 0.01 & 0.01 & 0.01 & 0.00 & 0.00 & 0.01 \\
 & Trans-EnV greatly preferred to \methodName & 0.00 & 0.00 & 0.00 & 0.01 & 0.00 & 0.00 \\
\midrule
 & \methodName greatly preferred to Trans-EnV & 0.93 & 0.99 & 0.69 & 0.81 & 0.99 & 0.62 \\
 & \methodName somewhat preferred to Trans-EnV & 0.07 & 0.01 & 0.30 & 0.15 & 0.01 & 0.36 \\
Model & \methodName approximately equal to Trans-EnV & 0.00 & 0.00 & 0.01 & 0.02 & 0.00 & 0.02 \\
 & Trans-EnV somewhat preferred to \methodName & 0.00 & 0.00 & 0.00 & 0.00 & 0.00 & 0.00 \\
 & Trans-EnV greatly preferred to \methodName & 0.00 & 0.00 & 0.00 & 0.02 & 0.00 & 0.00 \\
\bottomrule
\end{tabular}%
}
\end{table}

\subsection{Validating User--Model Splits: Human Preference Annotations}
\label{app:user_model_human_detail}
Similar to Sec. \ref{app:preference_annotations}, we present native speakers with 100 side-by-side pairs, each containing an DialectLLM-User and DialectLLM-Model. We ask the annotators which response better represents how a model should respond in their dialect in a 5-point scale. The order of the options was shuffled to ensure no annotation bias. The quantitative results are shown in Table \ref{tab:detailed_user_model_preference_distribution}, where we observe a strong preference (80.65\%) towards DialectLLM-Model for the model response, while the opposite direction holds for only 4.23\% of the cases. 

\begin{table*}[h]
\centering
\caption{
Detailed DialectLLM-Model vs DialectLLM-User preference distribution.
Native speakers compared 100 shuffled response pairs per dialect. 
}
\label{tab:detailed_user_model_preference_distribution}
\small
\setlength{\tabcolsep}{5pt}
\begin{tabular}{lcccccc}
\toprule
Preference & AU & IN & IE & NG & PH & SC \\
\midrule
DialectLLM-Model greatly preferred to DialectLLM-User  & 0.35 & 0.15 & 0.68 & 0.68 & 0.42 & 0.09  \\
DialectLLM-Model somewhat preferred to DialectLLM-User& 0.45 & 0.38 & 0.31 & 0.23 & 0.46 & 0.65 \\
DialectLLM-Model approximately equal to DialectLLM-User & 0.19 & 0.41 & 0.00  & 0.01  & 0.07  & 0.22 \\
DialectLLM-User somewhat preferred to DialectLLM-Model & 0.01  & 0.06  & 0.01  & 0.04  & 0.05  & 0.04  \\
DialectLLM-User greatly preferred to DialectLLM-Model  & 0.00  & 0.00  & 0.00  & 0.04  & 0.00  & 0.00  \\
\bottomrule
\end{tabular}

\end{table*}

Beyond aggregate preferences, annotator comments consistently reveal that DialectLLM-User often preserves dialect features that native speakers perceive as natural for user utterances but inappropriate for assistant/model responses. Representative comments are shown in Table~\ref{tab:user_model_qualitative_comments}. These comments closely align with our feature-level annotations (Sec. \ref{subsec:wordbank_construction}), where features such as discourse-marker \textit{like}, colloquial second-person pronouns, zero-article usage, and pluralized uncountables are marked as user-understandable but not model-appropriate. Thus, the qualitative feedback helps explain the strong preference for DialectLLM-Model and provides a validation of the user--model split.

\begin{table*}[h]
\centering
\caption{
Representative annotator comments for the DialectLLM-Model versus DialectLLM-User preference annotation. Display order was randomized during annotation; we replace references to the two variants for readability.}
\label{tab:user_model_qualitative_comments}
\footnotesize
\setlength{\tabcolsep}{5pt}
\renewcommand{\arraystretch}{1.15}
\begin{tabular}{p{0.08\textwidth}p{0.86\textwidth}}
\toprule
Dialect & Representative comments \\
\midrule
IN &
- DialectLLM-Model is greatly preferred because it avoids unnecessary filler words and maintains clear grammar suitable for an assistant response. \newline
- DialectLLM-Model is an excellent example of my dialect and is grammatically correct. \\

\midrule
IR &
- No issues with DialectLLM-Model, but DialectLLM-User is too colloquial. \newline
- DialectLLM-User contains examples that are not appropriate for written AI. \\

\midrule
SC &
- DialectLLM-Model is greatly preferred; DialectLLM-User is too informal
for the assistant response. \newline
- DialectLLM-User uses grammar that the model should not use. For example, the use of \textit{like}, \textit{ye}, or \textit{youse} is a behavior that the model should not copy. \\

\midrule
NG &
- DialectLLM-Model maintains a helpful and `neighbourly' tone while correctly using
the Nigerian spelling of \textit{organise}. While DialectLLM-User attempts dialectal
markers such as zero-article usage and pluralized uncountables, these result in an
over-transformed or overly informal tone that the assistant should generally avoid. \\
\bottomrule
\end{tabular}

\end{table*}
\clearpage
\section{LLM Benchmarking for Open-Ended Dialectal Generation}
\label{app:llm_generation}
Considering that the current LLMs are not well-equipped for dialectal capabilities, concrete evaluation of generation capabilities through LLMs in near impossible. Taking into consideration, we test the generation capabilities with a proxy task of response completion, where we give the candidate response to an LLM and prompt the model to choose the dialect-appropriate response as shown in Sec.~\ref{subsec:exp_response_completion}. However, to get a glimpse of the current models' capabilities truly on generation when guided to output a certain dialect, we evaluate the output results of LLMs with different LLM-as-a-Judges (LLMaaJs)~\cite{gu2024survey}. To do this, we for each target turn (e.g., turn 8), we provide the model with the preceding conversation history (turns 1 through 7) and the last turn's user utterance and prompt the model to respond in a specified dialect. For the generated output, we evaluate the models' performances with three LLMaaJs, using Claude-4-Sonnet), where we construct separate evaluators for each orthographic, lexical, and morphosyntactic features. Each judge is provided with human annotations as reference for their assessment and performs binary classification (valid or invalid), with an N/A option when the generated text contains no relevant markers.

LLMaaJ proves suboptimal for generation evaluation. This roots from two causes: (1) LLMs lack proper linguistic knowledge for dialects and (2) models exhibit high hallucination rates even on straightforward binary judgment~\cite{oh2024erbench}. Despite the limitation, we report LLMaaJ as a rough indicator for relative performance differences across different models. When computing the scores, we exclude the abstained (N/A) responses. This assessment introduces a slight advantage for smaller models, since when a model fails to produce coherent English entirely (e.g., outputting Tagalog when prompted for Philippine English or generating repetitive nonsense), which happens occasionally for smaller models, judges abstain rather than penalize. This likely explains the unexpected performance drop from Qwen3-14B to Qwen3-32B.
\begin{figure}[htbp]
    \centering
    \includegraphics[width=\linewidth]{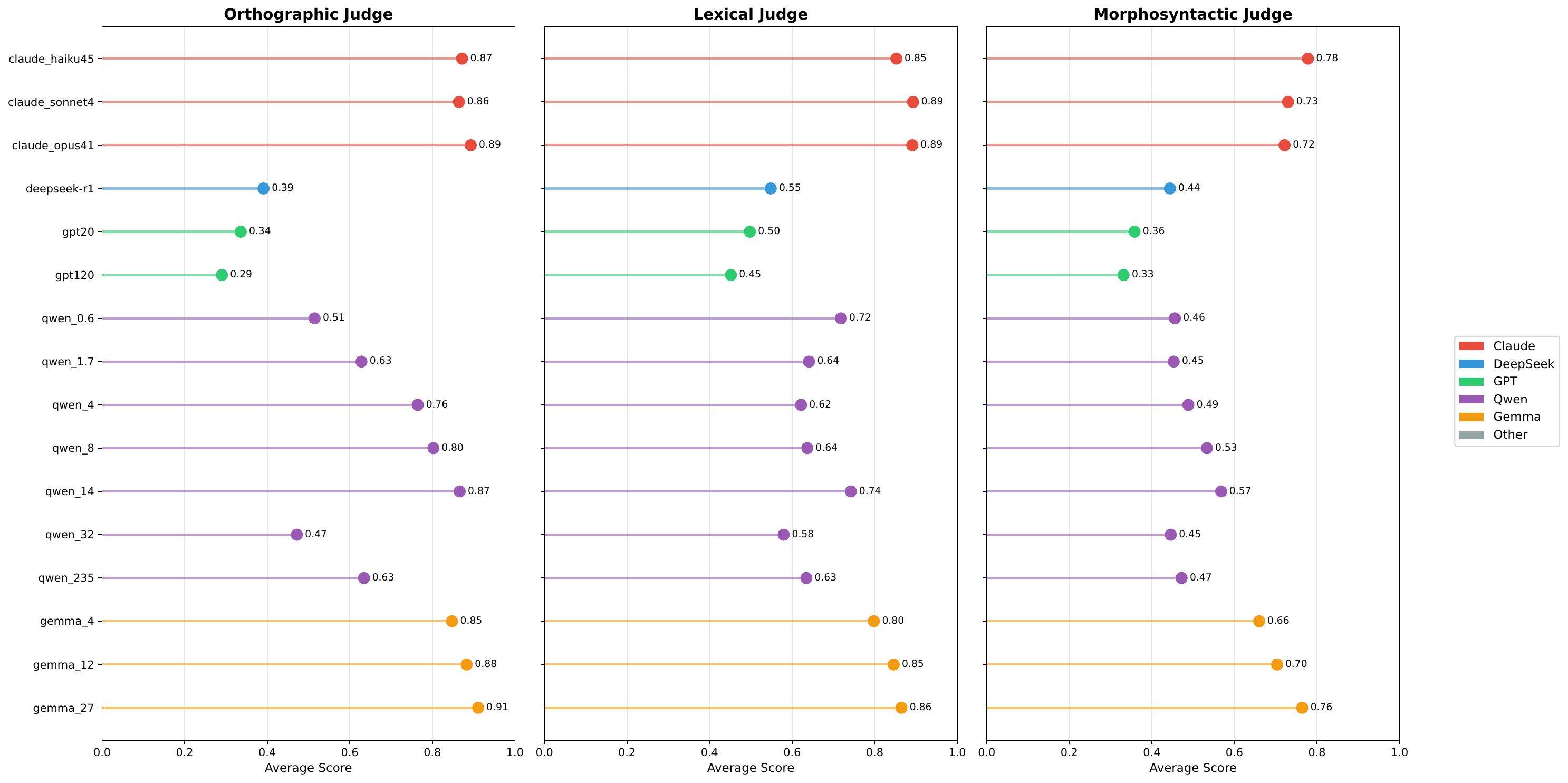}

    \caption{LLMaaJ evaluation results for open-ended generation. Scores across orthographic, lexical, and morphosyntactic judges confirm that performance scales with model size, consistent with other tasks.}

    \label{fig:generation_llmaaj}
\end{figure}

As shown in Fig.~\ref{fig:generation_llmaaj}, larger models within each model family generally outperform smaller variants, with Claude and Gemma demonstrating the strongest generation capabilities overall. Given LLMaaJ's limitations, these scores should be interpreted as relative comparisons, preferably within the model family, rather than absolute measures of dialect generation quality. Moreover, we can observe that LLMs are less capable of incorporating appropriate morphosyntactic features compared to lexical or orthographic features. We leave the development of generation metrics and dialectal post-training for better open-ended generation as important directions for future research.

\clearpage
\section{Additional Results}
\subsection{More Results from Sec.~\ref{subsec:main_results} }
\label{app:more_classification_generation}
\begin{figure*}[htbp]
    \centering
    \includegraphics[width=\linewidth]{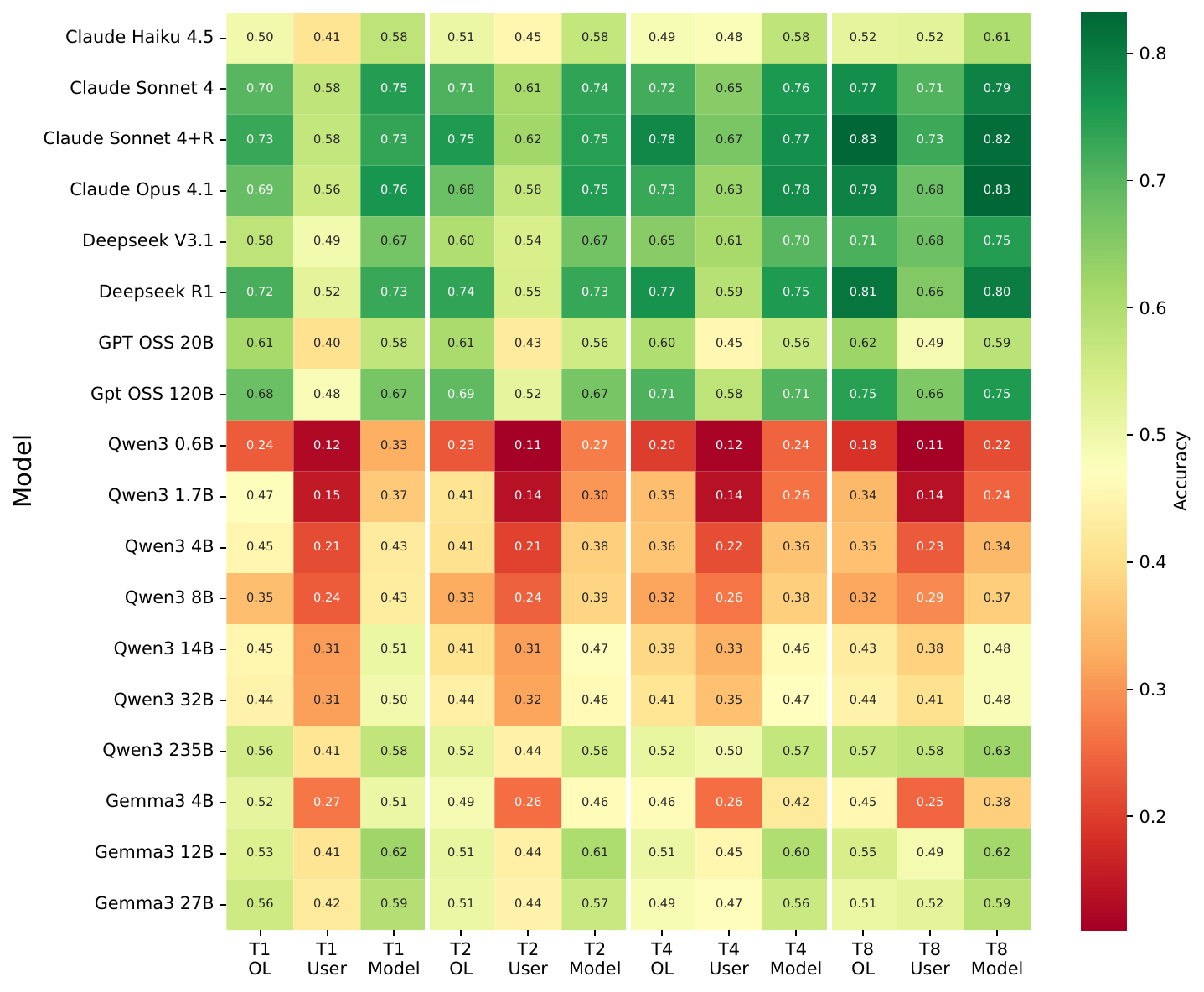}

    \caption{Model performance on \benchmarkName classification (Identification) task. We observe an overall trend of positive correlation between model size \& accuracy and that models struggle when morphosyntactic features are added. OL, User, and Model denote the DialectLLM-OL, -User, and -Model dataset, respectively.}

    \label{fig:classification}
\end{figure*}

\begin{figure}[htbp]
    \centering
    \includegraphics[width=\linewidth]{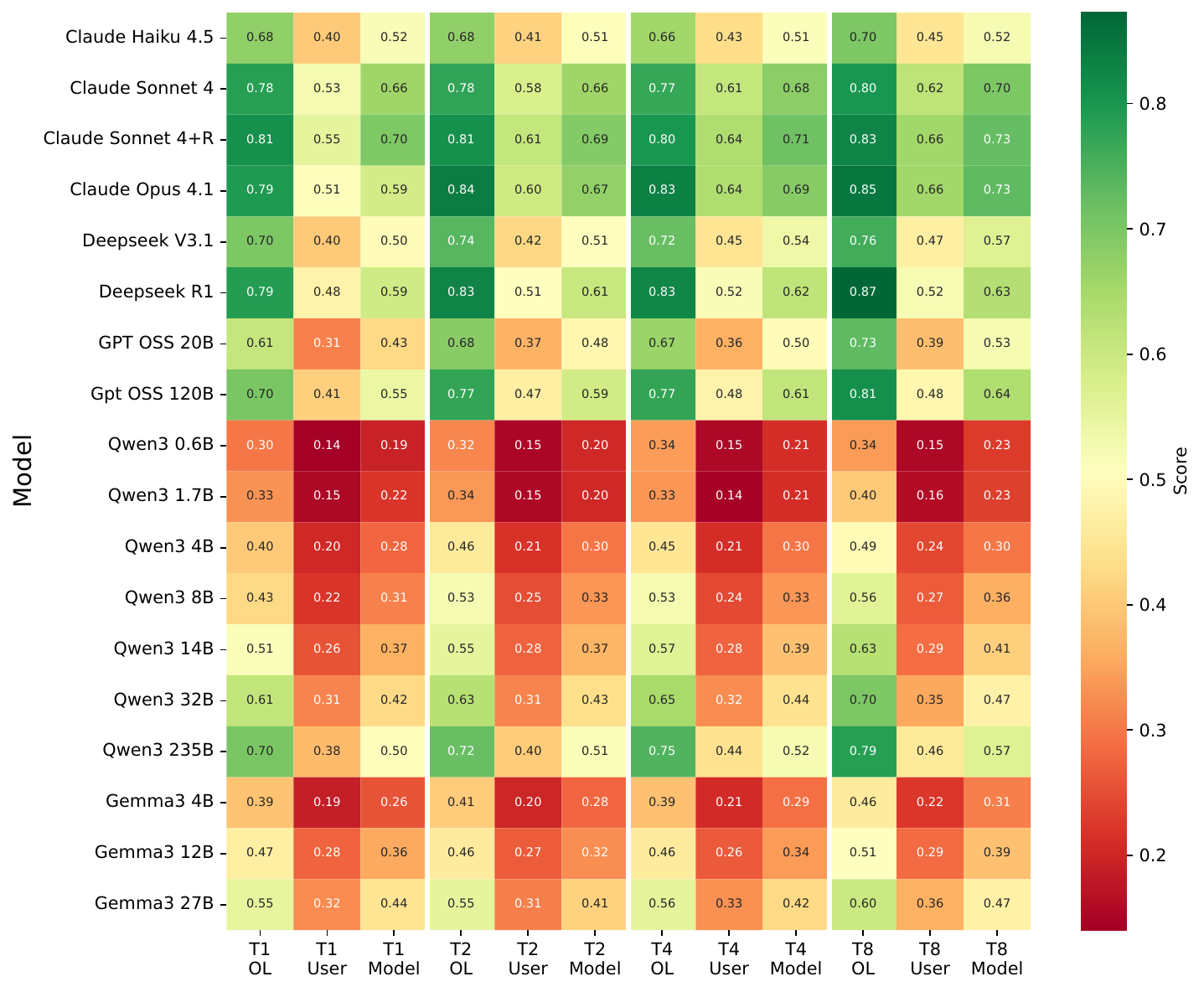}

    \caption{Model performance on \benchmarkName generation (response completion) task. Similar trend to Fig.~\ref{fig:classification} is observed.}

    \label{fig:response_completion}
\end{figure}

We present the full results for the classification (identification) and generation (response completion) task in Fig.~\ref{fig:classification} and Fig.~\ref{fig:response_completion} across 1, 2, 4, and 8 turn dialogs. We observe similar trends for both classification and generation tasks, while models show a slight degradation in performance for the generation task compared to that of classification. For classification, to provide context for the results, we establish three baselines for the classification task across 1, 2, 4, and 8 turns. A \textit{random guess} yields \{.31, .27, .24, .22\}, \textit{SAE-biased guess} (always selecting SAE as the answer) yields \{.21, .18, .17, .15\}, and \textit{GB-biased guess} yields \{.37, .33, .29, .26\} average accuracies, respectively.

\clearpage
\subsection{More Results from Sec.~\ref{subsec:model_size_performance}}
\label{app:model_size_perf}
We show the correlation between model size and performance for both classification and generation task on \benchmarkName. We see a linear positive correlation within the model family and also in general across families. 
\begin{figure*}[htb]
    \centering
    \includegraphics[width=0.95\linewidth]{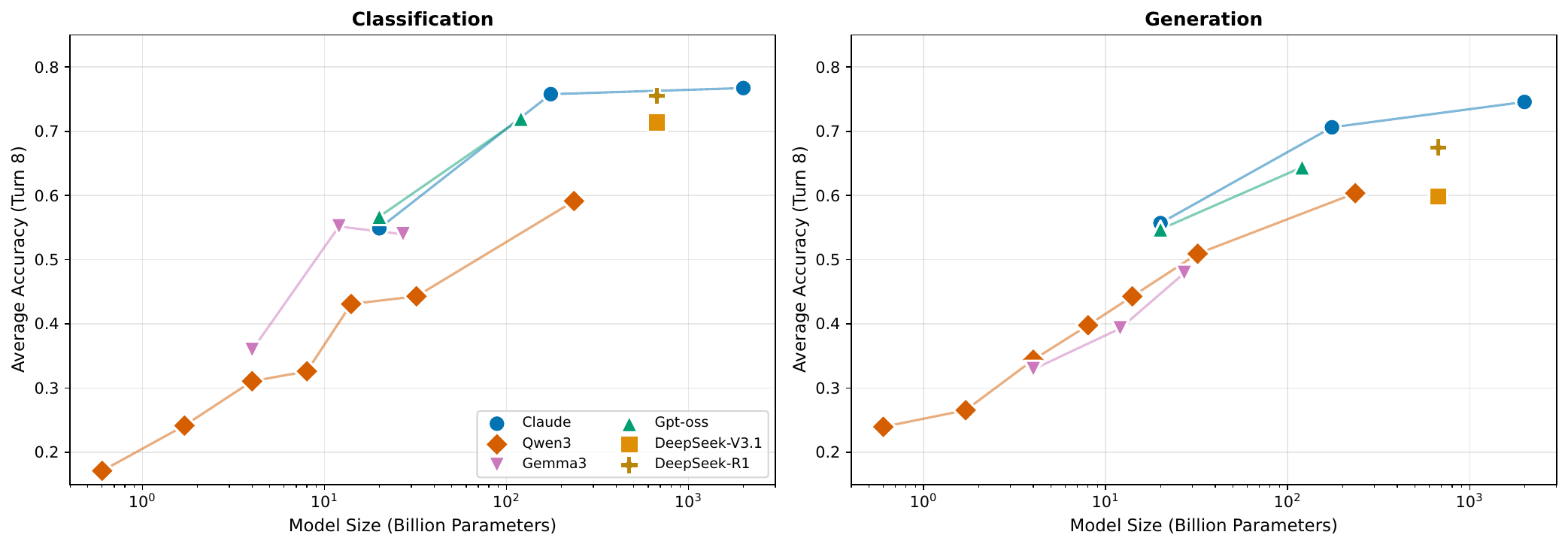}

    \caption{Scaling behavior of LLMs on dialectal classification and generation tasks. We show the model performance at turn 8, which has the most contextual cues and has the least multi-labels, to avoid giving credit for models outputting US or GB-biased answers. Accuracy scales linearly with model size.}

    \label{fig:model_size_performance}
\end{figure*}

\subsection{More Results from Sec.~\ref{subsec:turn_performance}}
\label{app:turn_perf}

\begin{figure}[h!]
     \centering
     % First Subfigure
     \begin{subfigure}[b]{0.49\textwidth}
         \centering
         \includegraphics[width=\textwidth]{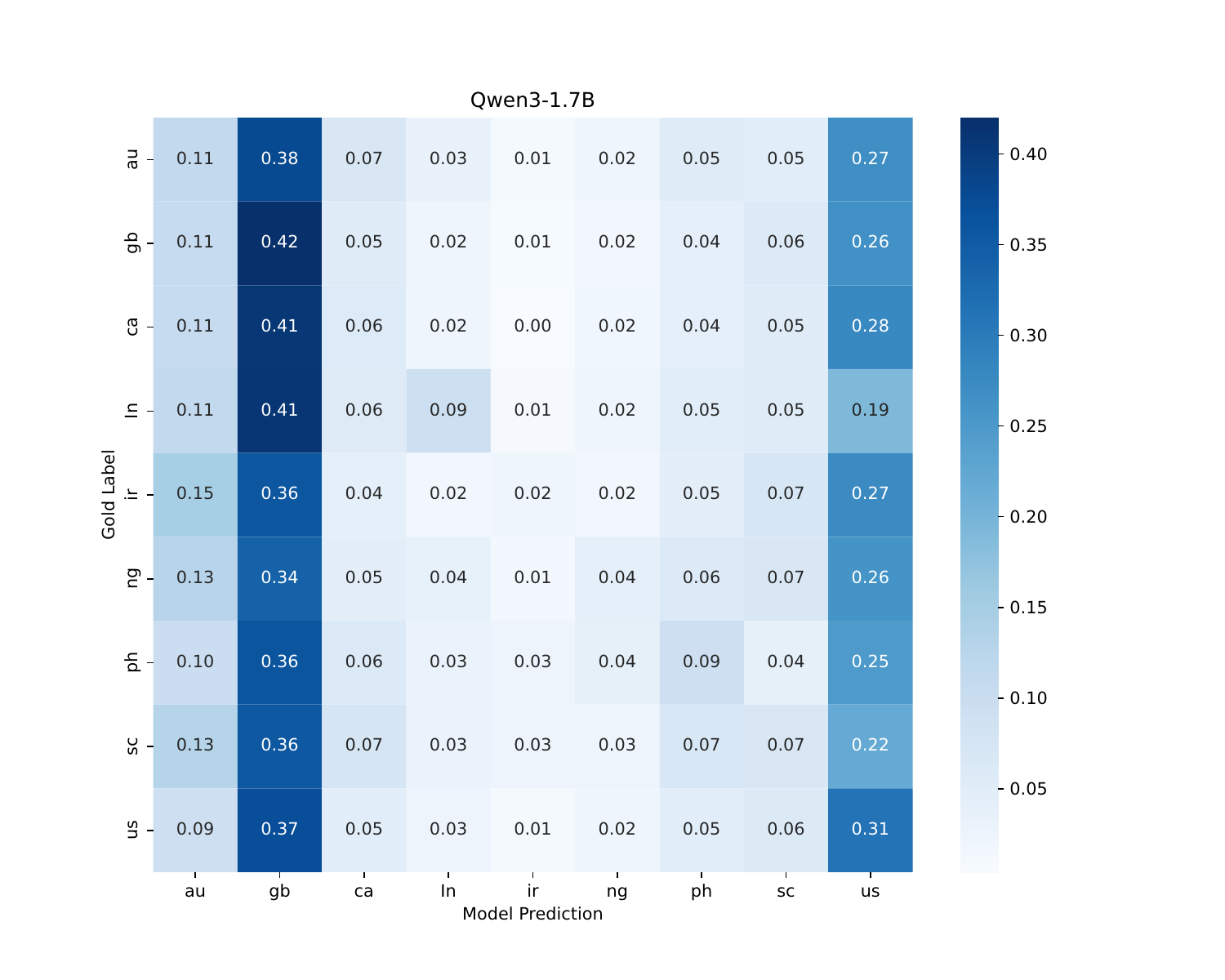}
     \end{subfigure}
     \hfill % This adds horizontal space between the images
     % Second Subfigure
     \begin{subfigure}[b]{0.49\textwidth}
         \centering
         \includegraphics[width=\textwidth]{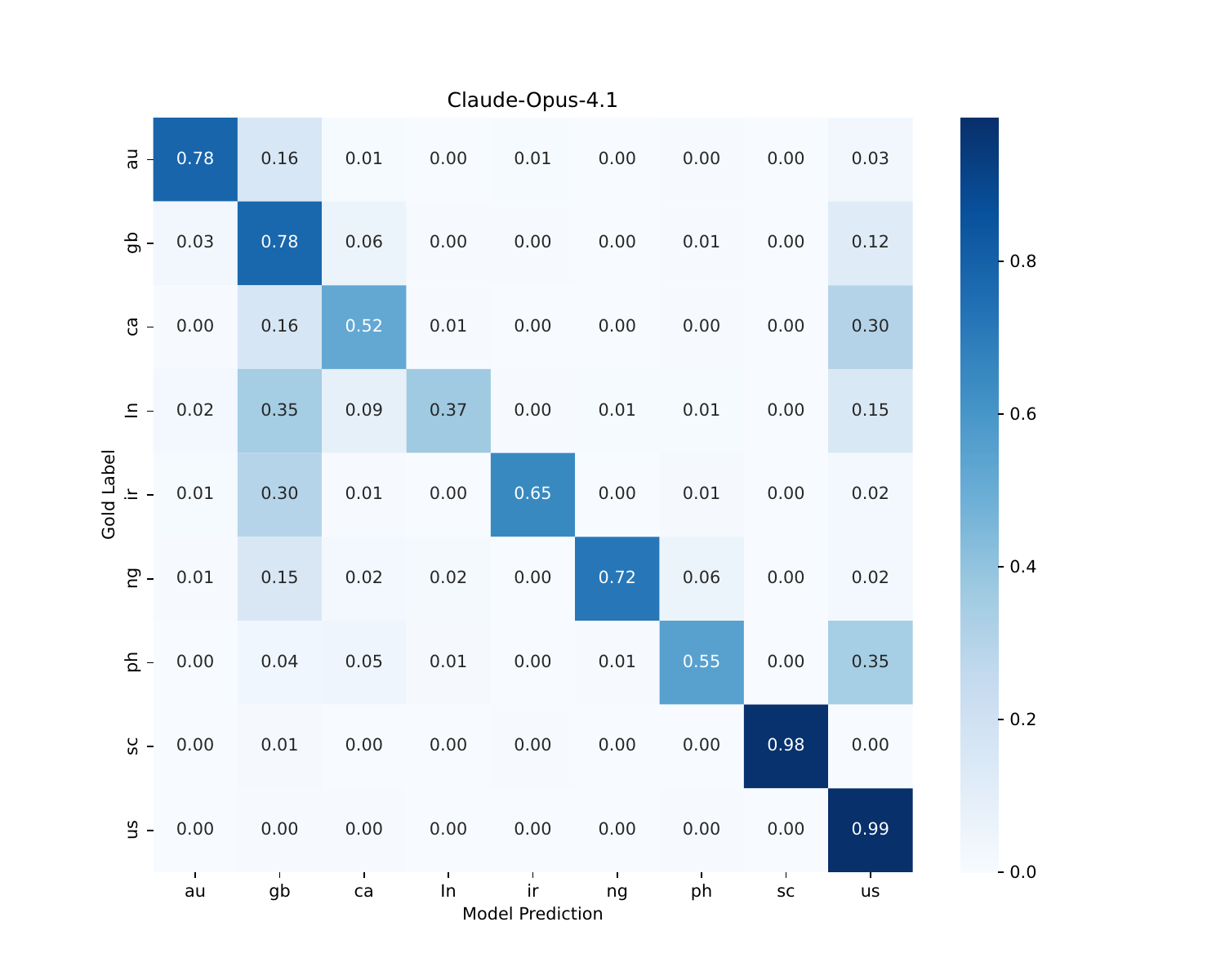}
     \end{subfigure}
     \caption{Confusion matrix for Qwen3-1.7B (left) and Claude-Opus-4.1 (right) on 8-turn dialogs; classification task. For Qwen3-1.7B, The vertical clustering in the US and GB columns suggests that the model is making default guesses to standard dialects rather than analyzing dialectal features. On the other hand, for Claude-Opus-4.1, the diagonal pattern indicates that the model is capable of genuinely identifying dialects.}
     \label{fig:qwen_claude_t8_confusion}
\end{figure}

\begin{figure}[h!]
    \centering
    \includegraphics[width=\linewidth]{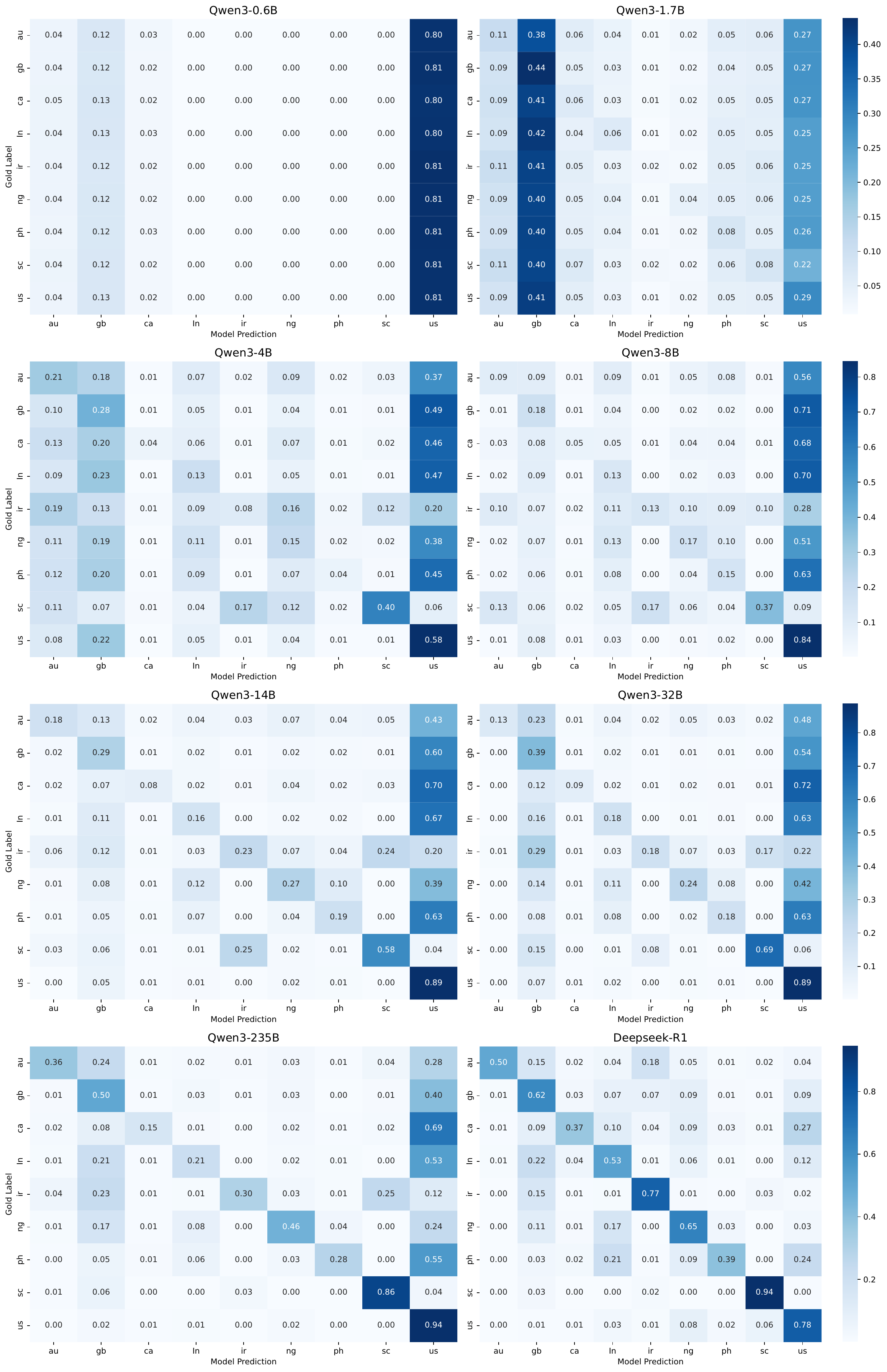}

    \caption{Confusion matrices of different models averaged over all turns (1, 2, 4, and 8); classification task. Smaller models show more bias towards high-resource, standard dialects (US, GB).}
    \label{fig:all_models_confusion_1}
   
\end{figure}

\begin{figure}[htbp]
    \centering
    \includegraphics[width=\linewidth]{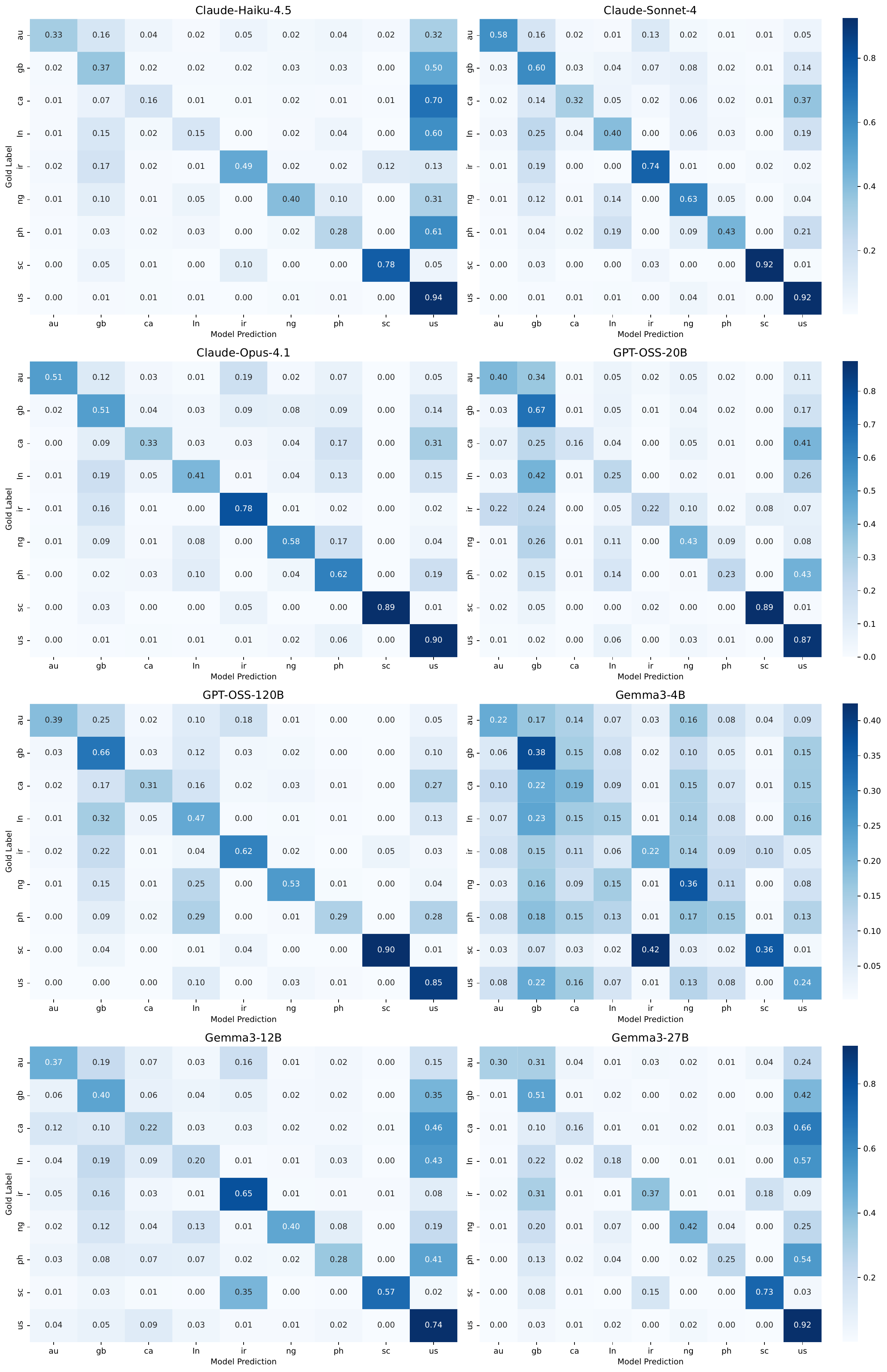}

    \caption{Confusion matrices of different models averaged over all turns (1, 2, 4, and 8); classification task. Smaller models show more bias towards high-resource, standard dialects (US, GB).}
    \label{fig:all_models_confusion_2}

\end{figure}

In this section, we show the confusion matrices on the classification task of different models. More context typically incorporates more dialectal cues, hence the likelihood of a unique gold label gets higher. Considering that a standard confusion matrix requires a single gold label, we report the confusion matrix for two representative models (Qwen3-1.7B and Claude-Opus-4.1) for 8-turn dialogs as a visual representation (Fig.~\ref{fig:qwen_claude_t8_confusion}). We also present the confusion matrices of different models aggregated across all turns and datasets in Fig.~\ref{fig:all_models_confusion_1} and Fig.~\ref{fig:all_models_confusion_2}.

\clearpage

\subsection{More Results from Sec.~\ref{subsec:post-training}}
\label{app:post-training}
\begin{table*}[ht]
\centering
\caption{Model performance results on \benchmarkName classification task after post-training. Models with suffix ``-0.15'' are models trained on 15\% of the training pool and those with suffix ``-full'' are trained on the whole training data. \texttt{OL}, \texttt{U}, and \texttt{M} denote DialectLLM-OL, -User, and -Model, respectively.}
\resizebox{\textwidth}{!}{
\begin{tabular}{l ccc ccc ccc ccc}
\toprule
 & \multicolumn{3}{c}{\textbf{Turn 1}} & \multicolumn{3}{c}{\textbf{Turn 2}} & \multicolumn{3}{c}{\textbf{Turn 4}} & \multicolumn{3}{c}{\textbf{Turn 8}} \\
\cmidrule(lr){2-4} \cmidrule(lr){5-7} \cmidrule(lr){8-10} \cmidrule(lr){11-13}
\textbf{Model} & \texttt{OL} & \texttt{U} & \texttt{M} & \texttt{OL} & \texttt{U} & \texttt{M} & \texttt{OL} & \texttt{U} & \texttt{M} & \texttt{OL} & \texttt{U} & \texttt{M} \\
\midrule
Qwen3-8B & 0.353 & 0.237 & 0.429 & 0.326 & 0.240 & 0.385 & 0.318 & 0.265 & 0.377 & 0.319 & 0.287 & 0.373 \\
Qwen3-8B-0.15 & 0.746 & 0.878 & 0.919 & 0.782 & 0.941 & 0.957 & 0.838 & 0.976 & 0.977 & 0.887 & 0.991 & 0.990 \\
Qwen3-8B-Full & 0.860 & 0.956 & 0.967 & 0.872 & 0.990 & 0.991 & 0.910 & 0.998 & 0.996 & 0.944 & 0.999 & 0.999 \\
Gemma3-12B          & 0.535 & 0.410 & 0.622 & 0.511 & 0.436 & 0.605 & 0.507 & 0.450 & 0.599 & 0.546 & 0.493 & 0.617 \\
Gemma3-12B-Full & 0.875 & 0.967 & 0.967 & 0.882 & 0.990 & 0.990 & 0.915 & 0.998 & 0.997 & 0.949 & 1.000 & 0.999 \\
\bottomrule
\end{tabular}
}
\end{table*}

\begin{table*}[ht]
\centering
\caption{Model performance results on \benchmarkName generation task  after post-training. Models with suffix ``-0.15'' are models trained on 15\% of the training pool and those with suffix ``-full'' are trained on the whole training data. \texttt{OL}, \texttt{U}, and \texttt{M} denote DialectLLM-OL, -User, and -Model, respectively.}
\resizebox{\textwidth}{!}{
\begin{tabular}{l ccc ccc ccc ccc}
\toprule
 & \multicolumn{3}{c}{\textbf{Turn 1}} & \multicolumn{3}{c}{\textbf{Turn 2}} & \multicolumn{3}{c}{\textbf{Turn 4}} & \multicolumn{3}{c}{\textbf{Turn 8}} \\
\cmidrule(lr){2-4} \cmidrule(lr){5-7} \cmidrule(lr){8-10} \cmidrule(lr){11-13}
\textbf{Model} & \texttt{OL} & \texttt{U} & \texttt{M} & \texttt{OL} & \texttt{U} & \texttt{M} & \texttt{OL} & \texttt{U} & \texttt{M} & \texttt{OL} & \texttt{U} & \texttt{M} \\
\midrule
Qwen3-8B & 0.426 & 0.220 & 0.310 & 0.525 & 0.250 & 0.334 & 0.527 & 0.243 & 0.333 & 0.563 & 0.268 & 0.363 \\
Qwen3-8B-0.15 & 0.896 & 0.900 & 0.952 & 0.915 & 0.887 & 0.932 & 0.920 & 0.888 & 0.923 & 0.944 & 0.888 & 0.926 \\
Qwen3-8B-Full & 0.935 & 0.972 & 0.985 & 0.948 & 0.955 & 0.962 & 0.947 & 0.952 & 0.961 & 0.961 & 0.949 & 0.961 \\
Gemma3-12B & 0.468 & 0.275 & 0.358 & 0.459 & 0.267 & 0.324 & 0.458 & 0.264 & 0.343 & 0.507 & 0.285 & 0.392 \\
Gemma3-12B-Full & 0.946 & 0.978 & 0.988 & 0.953 & 0.963 & 0.971 & 0.950 & 0.958 & 0.967 & 0.962 & 0.954 & 0.967 \\
\bottomrule
\end{tabular}
}
\end{table*}

\subsection{More Results from Sec.~\ref{subsec:cross-task}}
\label{app:cross-task}

\begin{table}[h!]
\centering
\caption{Model performance on \benchmarkName classification task, which are fine-tuned on different tasks. ``-RC'' denotes that the model is trained on the response completion (generation) task and ``-Id'' denotes that the model is trained on the identification (classification) task. \texttt{OL}, \texttt{U}, and \texttt{M} denote DialectLLM-OL, -User, and -Model, respectively.}
\resizebox{\textwidth}{!}{
\begin{tabular}{l ccc ccc ccc ccc}
\toprule
 & \multicolumn{3}{c}{\textbf{Turn 1}} & \multicolumn{3}{c}{\textbf{Turn 2}} & \multicolumn{3}{c}{\textbf{Turn 4}} & \multicolumn{3}{c}{\textbf{Turn 8}} \\
\cmidrule(lr){2-4} \cmidrule(lr){5-7} \cmidrule(lr){8-10} \cmidrule(lr){11-13}
\textbf{Model} & \texttt{OL} & \texttt{U} & \texttt{M} & \texttt{OL} & \texttt{U} & \texttt{M} & \texttt{OL} & \texttt{U} & \texttt{M} & \texttt{OL} & \texttt{U} & \texttt{M} \\
\midrule
Qwen3-8B          & 0.353 & 0.237 & 0.429 & 0.326 & 0.240 & 0.385 & 0.318 & 0.265 & 0.377 & 0.319 & 0.287 & 0.373 \\
Qwen3-8B-RC-0.15        & 0.548 & 0.425 & 0.589 & 0.553 & 0.395 & 0.518 & 0.491 & 0.306 & 0.452 & 0.484 & 0.357 & 0.448 \\
Qwen3-8B-RC       & 0.523 & 0.346 & 0.511 & 0.504 & 0.325 & 0.475 & 0.438 & 0.293 & 0.428 & 0.394 & 0.241 & 0.383 \\
Qwen3-8B-Id-0.15        & 0.803 & 0.855 & 0.901 & 0.811 & 0.911 & 0.939 & 0.856 & 0.957 & 0.968 & 0.901 & 0.983 & 0.987 \\
Qwen3-8B-Id & 0.863 & 0.952 & 0.963 & 0.872 & 0.983 & 0.985 & 0.908 & 0.995 & 0.994 & 0.942 & 0.999 & 0.999 \\
Qwen3-8B-Both      & 0.860 & 0.956 & 0.967 & 0.872 & 0.990 & 0.991 & 0.910 & 0.998 & 0.996 & 0.944 & 0.999 & 0.999 \\
\bottomrule
\end{tabular}
}
\label{tab:app_cross_identification}
\end{table}
\begin{table}[h!]
\centering
\caption{Model performance on \benchmarkName generation task, which are fine-tuned on different tasks. ``-RC'' denotes that the model is trained on the response completion task and ``-Id'' denotes that the model is trained on the identification task. Models with suffix ``-0.15'' are models trained on 15\% of the training pool. \texttt{OL}, \texttt{U}, and \texttt{M} denote DialectLLM-OL, -User, and -Model, respectively.}
\resizebox{\textwidth}{!}{
\begin{tabular}{l ccc ccc ccc ccc}
\toprule
 & \multicolumn{3}{c}{\textbf{Turn 1}} & \multicolumn{3}{c}{\textbf{Turn 2}} & \multicolumn{3}{c}{\textbf{Turn 4}} & \multicolumn{3}{c}{\textbf{Turn 8}} \\
\cmidrule(lr){2-4} \cmidrule(lr){5-7} \cmidrule(lr){8-10} \cmidrule(lr){11-13}
\textbf{Model} & \texttt{OL} & \texttt{U} & \texttt{M} & \texttt{OL} & \texttt{U} & \texttt{M} & \texttt{OL} & \texttt{U} & \texttt{M} & \texttt{OL} & \texttt{U} & \texttt{M} \\
\midrule
Qwen3-8B             & 0.426 & 0.220 & 0.310 & 0.525 & 0.250 & 0.334 & 0.527 & 0.243 & 0.333 & 0.563 & 0.268 & 0.363 \\
Qwen3-8B-RC-0.15        & 0.884 & 0.881& 0.940 & 0.915 & 0.871 & 0.917 & 0.911 & 0.884 & 0.912 & 0.944 & 0.880 & 0.909 \\
Qwen3-8B-RC    & 0.930 & 0.963 & 0.982 & 0.944 & 0.945 & 0.962 & 0.946 & 0.946 & 0.959 & 0.963 & 0.946 & 0.961 \\
Qwen3-8B-Id-0.15        & 0.298 & 0.199 & 0.243 & 0.318 & 0.221 & 0.238 & 0.328 & 0.223 & 0.246 & 0.347 & 0.215 & 0.212 \\
Qwen3-8B-Id    & 0.173 & 0.114 & 0.127 & 0.244 & 0.170 & 0.184 & 0.292 & 0.227 & 0.237 & 0.291 & 0.226 & 0.245 \\
Qwen3-8B-Both         & 0.935 & 0.972 & 0.985 & 0.948 & 0.955 & 0.962 & 0.947 & 0.952 & 0.961 & 0.961 & 0.949 & 0.961 \\
\bottomrule
\end{tabular}
}
\label{tab:app_cross_rc}
\end{table}

\clearpage
\subsection{Impact of Inference-Time Reasoning on \benchmarkName}
\label{subsec:reasoning}
\begin{figure}[h!]
     \centering
     % First Subfigure
     \begin{subfigure}[b]{0.49\textwidth}
         \centering
         \includegraphics[width=\textwidth]{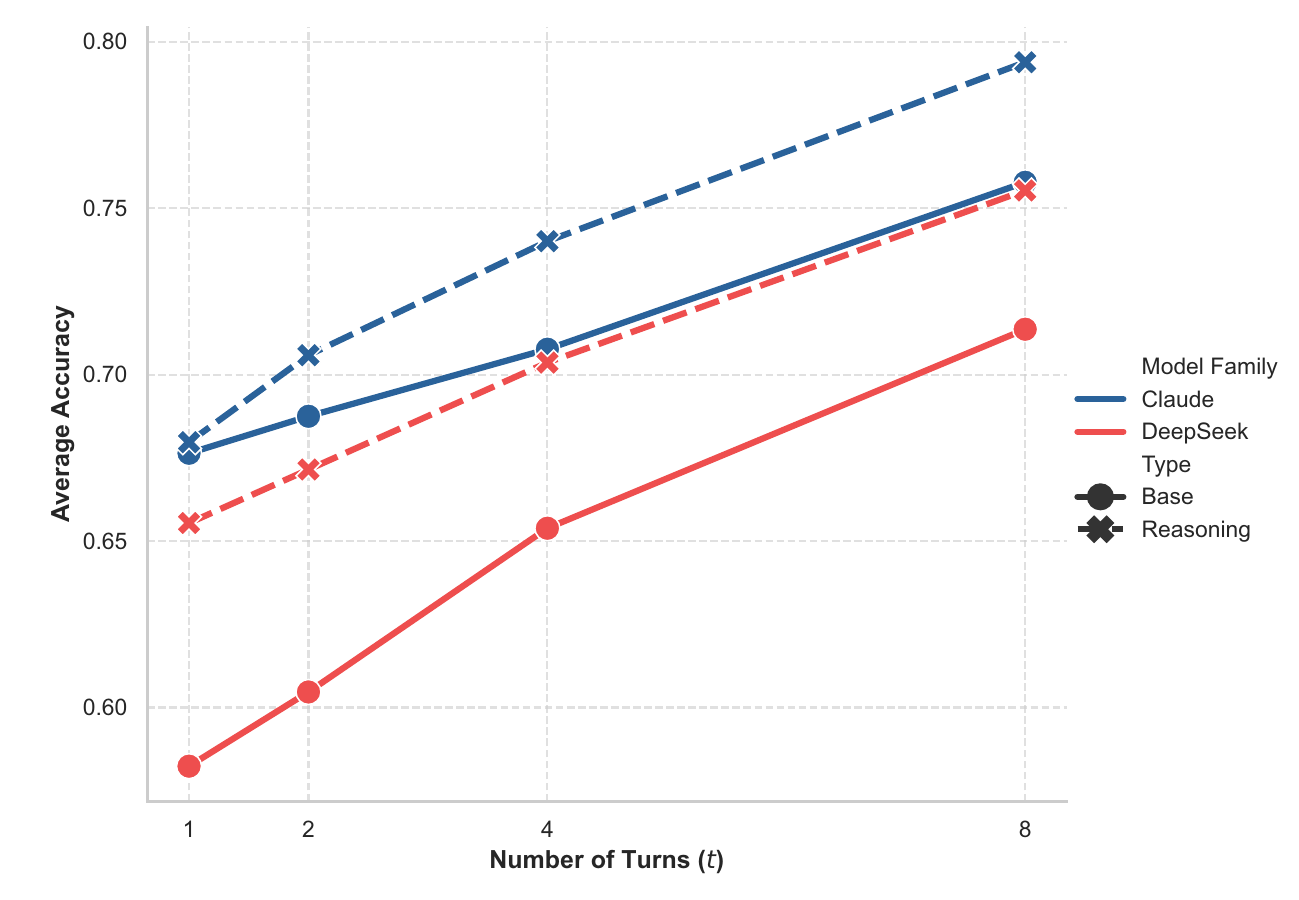}
         \label{fig:first_reasoning}
     \end{subfigure}
     \hfill % This adds horizontal space between the images
     % Second Subfigure
     \begin{subfigure}[b]{0.49\textwidth}
         \centering
         \includegraphics[width=\textwidth]{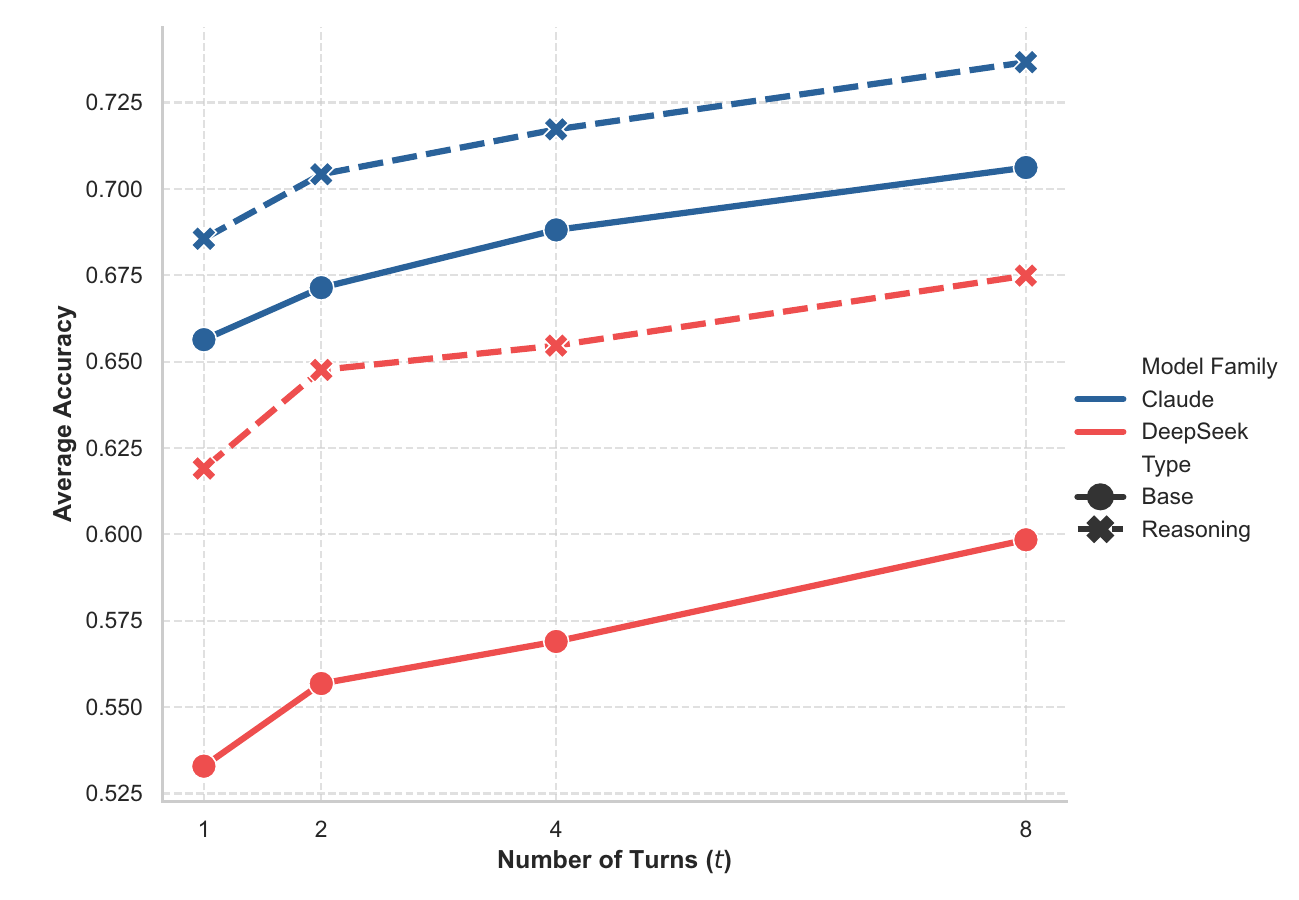}
         \label{fig:second_reasoning}
     \end{subfigure}
     \caption{Impact of test-time compute on classification (left) and generation (right) task. Models equipped with reasoning capabilities (dotted lines) achieve a consistent performance increase over their base counterparts.}
     \label{fig:reasoning_response_completion}
\end{figure}

We further investigate whether utilizing test-time compute enhances LLM performance on \benchmarkName, comparing Claude-Sonnet-4 (with and without thinking mode) and DeepSeek-V3.1 versus DeepSeek-R1. As illustrated in Fig.~\ref{fig:reasoning_response_completion}, inference-time reasoning yields a substantial and consistent accuracy gain across all turns, indicating that additional computational overhead enables synthesizing subtle dialectal cues.

\subsection{Cross-Variant Transfer}
\label{app:cross-variant}

Continuing from Sec.~\ref{subsec:generalization_data_distribution}, as a controlled proxy for distribution shift, we construct training and held-out test sets from each DialectLLM variant (-OL, -User, and -Model) using DialectLLM-Bench task formats. We train Qwen3-8B on each DialectLLM variant separately and evaluate it on all three variants. 
\begin{table}[h]
\centering
\caption{Cross-variant transfer results for Qwen3-8B. Each cell reports dialect identification / response completion
accuracy (\%). Bold entries indicate matched training and evaluation variants.}
\small
\begin{tabular}{lccc}
\toprule
Train $\downarrow$ / Test $\rightarrow$
& OL & User & Model \\
\midrule
Base Model
& 32.7 / 51.1
& 25.6 / 24.2
& 39.2 / 32.8 \\
OL
& \textbf{81.8 / 89.9}
& 58.0 / 63.2
& 72.9 / 71.4 \\
User
& 68.7 / 88.2
& \textbf{89.3 / 87.3}
& 85.7 / 84.7 \\
Model
& 67.1 / 88.3
& 48.2 / 71.2
& \textbf{92.6 / 91.6} \\
\bottomrule
\end{tabular}

\label{tab:cross_variant}
\end{table}

As shown in Table \ref{tab:cross_variant}, every cross-variant result exceeds the base model, indicating transfer across dialect realizations rather than fit to one dataset variant. Averaged over the six off-diagonal settings, identification accuracy rises from 32.5\% to 66.8\% (+34.3pp) and response completion accuracy from 36.0\% to 77.8\% (+41.8pp). These results suggest that the learned dialect competence is not tied to a particular data source.

\subsection{Evaluation on Naturally Occurring Data: International Corpus of English (ICE)}
\label{app:ice_data}
Evaluation on naturally occurring conversations provides complementary evidence about out-of-distribution generalization. Naturally occurring corpora differ across dialects in topic, setting, author background, and semantic content. Consequently, differences in model accuracy can be difficult to attribute specifically to dialect, as models could be picking up explicit context, such as physical locations and currencies, or implicit context, such as beliefs and values. DialectLLM holds the underlying dialog content constant, systematically varying native-validated lexical, orthographic, and morphosyntactic features, enabling more direct attribution of performance differences to dialectal variation.

\noindent\textbf{Experimental Setup.} We collect text from ICE, a corpus of written and spoken English segmented by nationality. We test whether our post-trained model (Qwen3-8B; 0.15 epoch), trained on all nine dialects, outperforms the baseline model on identifying the six overlapping dialects, AU, NG, IN, IE, PH, and US. Among the ICE data, we use conversation transcripts, for AU, NG, IN, IE, and PH, which often had its dialectal features embedded compared to formal written text. For US, the dataset contains only written registers, hence we use 30 personal letters and treat their numbered text units as pseudo-turns.

We filter our text with clear location or cultural signals such as ``Sydney'' for AU and ``Naira'' for NG, insufficient context (less than eight content words per turn), or text with no dialectal markers. The resulting evaluation set contains 4,761 dialogs across the 1-, 2-, 4-, and 8-turn settings. Following our benchmark design, we present the model with nine multiple-choice options (US, AU, GB, CA, IN, IE, NG, PH, and SC) and prompt them to choose the correct dialect.

\begin{table}[h]
\centering
\caption{Dialect identification accuracy (\%) of base and
post-trained Qwen3-8B on the filtered ICE evaluation set.
The six-way setting excludes GB, CA, SC from the candidate labels, which are absent from our evaluation set.
Bold indicates the higher accuracy within each setting.}
\small
\begin{tabular}{lrrrr}
\toprule
& \multicolumn{2}{c}{9-way}
& \multicolumn{2}{c}{6-way} \\
\cmidrule(lr){2-3}\cmidrule(lr){4-5}
Dialect & Base & Post-trained & Base & Post-trained \\
\midrule
Australian (AU) & 11.2 & \textbf{28.2} & 12.7 & \textbf{46.5} \\
Nigerian (NG)   & 27.3 & \textbf{55.3} & 32.1 & \textbf{58.2} \\
Indian (IN)     & 44.1 & \textbf{89.7} & 51.1 & \textbf{96.9} \\
Irish (IE)      & \textbf{31.2} & 29.2 & 48.2 & \textbf{75.8} \\
Philippine (PH) & 57.3 & \textbf{70.6} & 55.9 & \textbf{76.1} \\
American (US)   & 62.5 & \textbf{76.6} & \textbf{84.4} & 82.8 \\
\bottomrule
\end{tabular}
\label{tab:ice_transfer}
\end{table}

\noindent\textbf{Results.} As shown in Table \ref{tab:ice_transfer}, our post-trained model outperforms the base model in most of the cases, with gains ranging from 13.3 to 45.6 pp. The largest improvement is observed for IN (44.1\% $\rightarrow$ 89.7\%), followed by Nigerian English (27.3\% $\rightarrow$ 55.3\%). Irish English is the only exception, with accuracy decreasing slightly from 31.2\% to 29.2\%, where we analyze the cause in the following paragraph. These performance gains demonstrate that training on DialectLLM data enables the model to better identify dialects in this naturally occurring text.

We observe that the decrease for Irish English often comes from the confusion involving the phrase ``wee'', which is used in both Irish and Scottish English. The post-trained model frequently predicts dialogs containing ``wee'' as Scottish, while they are labeled as Irish based on the ICE dataset (no Scottish text in ICE). Dialect markers are often shared between dialects, so we cannot say that these texts are uniquely identifying. For instance, our post-trained model labels 49\% of AU dialogs as British and 59\% of IE dialogs as Scottish (base: 8\% and 29\%, respectively), which may be correct given the shared linguisitc features between these varieties. Across all dialects, 64\% of the post-trained model's errors fall on a historically related variety (GB for AU, NG, and IN; GB or SC for IE; US for PH; CA for US), compared with 13\% for the base model.

As an additional analysis, we remove dialogs, where ``wee'' is the only selected Irish marker. In this case, the post-trained model far exceeds the performance of the base model (58.7\% vs 33.2\%), consistent with improvements observed for other dialects. Moreover, we restrict the candidate labels to the six dialects existent in the ICE evaluation set. Under this setting (right two columns on Table \ref{tab:ice_transfer}), post-training improves average accuracy from 47.4\% to 72.7\%, IE accuracy from 48.2\% to 75.8\%, and AU accuracy from 12.7\% to 46.5\%.These results further support transfer from DialectLLM to naturally occurring text, suggesting that the model learns dialectal cues, while single-label evaluation may underestimate its performance when multiple dialect labels are plausible. This ambiguity is difficult to resolve in naturally occurring text, highlighting the value of DialectLLM’s controlled parallel data with the multi-labels (Sec. \ref{subsec:multi_label_assignment}).

\clearpage
\section{Models, Database, and Hyperparameter Details}
\label{app:model_hyperparameter}
\subsection{Models and Database}
\label{app:models_database_details}
We run Claude-Haiku-4.5, Claude-Sonnet-4, Claude-Opus-4.1, Qwen3-32B, Qwen3-235B-A22B, Deepseek-V3.1, Deepseek-R1, GPT-OSS-20B, and GPT-OSS-120B on Amazon Bedrock. We run Qwen3-\{0.6B, 1.7B, 4B, 8B, 14B\} (Apache-2.0) and Gemma3-\{4B, 12B, 27B\} (gemma) from Huggingface on 8 H100 or 8 A100 GPUs for inference and training. We use the rules from eWAVE database \cite{ewave} (CC-BY-3.0) for the morphosyntactic rules. 
\subsection{Hyperparameters}
\label{app:hyperparameters}
We present the hyperparameters used for supervised fine-tuning on classification and generation tasks. 
\begin{itemize}

    \item Learning rate = 2e-05

    \item Epoch = 1

    \item Learning Rate Scheduler= Cosine Annealing LR scheduler
 
    \item Maximum Gradient Norm = 1

    \item Warmup Ratio = 0.1
 
    \item Optimizer = AdamW
\end{itemize}

Moreover, for LLM inference calls on \benchmarkName, we set the temperature to 0 for non-reasoning models, where supported, to control randomness. For reasoning models, where setting the temperature to 0 is not supported, we did not specify any inference hyperparameters to minimize intervention.

\clearpage

\section{Prompts}
\label{app:prompts}
\subsection{Prompt for Sec.~\ref{subsec:seed_generation}}
\begin{figure}[h!]
    \centering
\begin{promptbox}{Seed Dialog Generation (Natural)}
        \sffamily % Use sans-serif for the instructions
        You are tasked with creating a natural, multi-turn conversation between a USER and an ASSISTANT about the topic: \texttt{(\{us\_word\})}

        \medskip
        \textbf{Rules:}
        \begin{itemize}[leftmargin=1.5em, nosep] % Requires \usepackage{enumitem}
            \item Generate a conversation in Standard American English (USA) with \texttt{\{num\_of\_turns\}} turns.
            \item If the topic \texttt{(\{us\_word\})} is related to spelling (e.g., words ending in -or), come up with an example that has the corresponding spelling structure.
            \item Make the conversation feel natural and conversational.
            \item Each turn = one USER message + one ASSISTANT response.
        \end{itemize}

        \medskip
        \textbf{Format:}
        
        \texttt{Turn 1:} \\
        \texttt{USER: [message]} \\
        \texttt{ASSISTANT: [response]}
        
        \medskip
        \texttt{Turn 2:} \\
        \texttt{USER: [message]} \\
        \texttt{ASSISTANT: [response]} \\
        \texttt{[continue as needed]}
    \end{promptbox}

    \caption{Prompt for \textit{natural} multi-turn seed dialog generation.}
\end{figure}

\begin{figure}[h!]
    \centering
    \begin{promptbox}{Seed Dialog Generation (Indirect)}
        \sffamily
        You are tasked with creating a natural, multi-turn conversation between a USER and an ASSISTANT about the topic: \texttt{[\{us\_word\}]}

        \vspace{0.8em}
        \textbf{Rules:}
        \begin{itemize}[leftmargin=1.5em, nosep]
            \item Generate a conversation in Standard American English (USA) with \texttt{\{num\_of\_turns\}} turns.
            \item If the topic \texttt{(\{us\_word\})} is related to spelling (e.g., words ending in -or), come up with an example that has the corresponding spelling structure.
            \item The USER must \textbf{not} directly mention the word \texttt{``\{us\_word\}''} in the \textbf{FIRST} turn.
            \item In the first turn, the USER should approach the topic indirectly through:
            \begin{itemize}[label=--, leftmargin=1.5em]
                \item Describing situations where it's needed
                \item Asking about related problems or scenarios
                \item Mentioning associated activities 
                \item Describing its features without naming it
            \end{itemize}
            \item Make the conversation feel natural and conversational.
            \item The ASSISTANT can mention \texttt{``\{us\_word\}''} in any turn. The USER can mention it starting from the \textbf{second turn}.
            \item Each turn = one USER message + one ASSISTANT response.
        \end{itemize}

        \vspace{0.8em}
        \textbf{Format:}

        \texttt{Turn 1:} \\
        \texttt{USER: [message]} \\
        \texttt{ASSISTANT: [response]}
        
        \vspace{0.4em}
        \texttt{Turn 2:} \\
        \texttt{USER: [message]} \\
        \texttt{ASSISTANT: [response]} \\
        \texttt{[continue as needed]}
    \end{promptbox}

    \caption{Prompt for \textit{indirect} multi-turn seed dialog generation.}
\end{figure}

\clearpage
\subsection{Prompt for Sec.~\ref{subsec:dialog_transform}}
\begin{figure}[h!]
    \centering
    \begin{promptbox}{Initial Orthographic and Lexical Transformation}
        \sffamily
        Convert the multi-turn dialogue from Standard American English Dialect to \texttt{\{target\_dialect\}} English Dialect.

        \vspace{0.8em}
        \textbf{Instructions:}
        \begin{itemize}[leftmargin=1.5em, nosep]
            \item Convert the dialogue while maintaining the original meaning, tone, and conversational flow
            \item Convert all measurements to \texttt{\{target\_dialect\}} standard units 
            \item Ensure the conversion reflects authentic linguistic features of the \texttt{\{target\_dialect\}} English dialect
            \item Focus on genuine dialectal variations rather than exaggerated stereotypical representations
            \item After the conversion, provide a detailed explanation of all changes made
        \end{itemize}

        \vspace{0.8em}
        \textbf{CONSTRAINTS:}
        \begin{itemize}[leftmargin=1.5em, nosep]
            \item Do not change the original morphosyntactic features. (differences in grammar, including how words are formed (morphology) and how they are arranged in sentences (syntax).)
            \item You are only allowed to make lexical, orthographic, and unit transformations mentioned below.
            \item Keep the \textbf{number and segmentation of turns} identical to the original.
        \end{itemize}

        \vspace{0.8em}
        \textbf{Dimensions to Consider:}
        \begin{itemize}[label=--, leftmargin=1.5em, nosep]
            \item Lexical Transformation: This dimension refers to the differences in vocabulary or the specific words used to refer to the same concept
            \item Orthographic Transformation: This dimension involves variations in the spelling of words. The pronunciation and meaning are generally the same, but the written form differs.
            \item Unit conversions (e.g., standard metric vs imperial system)
        \end{itemize}
        
        \smallskip
        \textit{\texttt{The \{target\_dialect\} English equivalent of ``\{us\_word\}'' is ``\{target\_word\}''.}}

        \vspace{0.8em}
        \textbf{FORMAT:} \\
        \texttt{CONVERTED DIALOGUE:} \\
        \texttt{[\{target\_dialect]\} English version here with grammar/prepositions/modals unchanged; same number of turns]} \\
        \texttt{[Use the Turn X format with USER/ASSISTANT labels as the original dialogue]}

        \medskip
        \texttt{EXPLANATION OF CHANGES:} \\
        \texttt{- List every change you made, grouped by dimension (Orthographic, Lexical).} \\
        \texttt{- Include an ``INTENTIONAL NON-CHANGES (Morphosyntax)'' section noting items you did not change because it would have altered grammar, prepositions, or modals.}

        \medskip
        \textbf{Original Dialogue to Convert:} \\
        \texttt{\{dialogue\}}
    \end{promptbox}

    \caption{Prompt template for the construction of the initial version of the DialectLLM-OL dataset (before refinement). The italicized sentence is included only when the mapping originates from the target transformation wordbank. }
\end{figure}
\clearpage
\subsection{Prompt for Sec~\ref{subsec:dialog_transform}}

\begin{figure}[ht]
    \centering
    \begin{promptbox}{Morphosyntactic Transformation}
        \sffamily
        \texttt{Previous Dialogue: \{prev\_dialogue\}}

        \vspace{0.8em}
        You are a linguistic expert specializing in \texttt{\{target\_dialect\}} English dialects. 
        \vspace{0.8em}
        Your task is to convert the text into an authentic-sounding \texttt{\{target\_dialect\}} English dialect based on the given rule and then explain the specific changes you made.

        \vspace{0.8em}
        \textbf{Instructions:}
        \begin{itemize}[leftmargin=1.5em, nosep]
            \item Refer to the previous dialogue for context understanding, do not transform the previous dialogue. 
            \item Your primary goal is to maintain a natural flow and preserve the original meaning. Apply the transformation rule listed in the ``Dialect Rule'' section.
            \item \textbf{Pay careful attention to the human comments for specific implementation details.}
            \item After the conversion, provide a detailed explanation of all changes made.
            \item Consider the previous dialogue context to ensure consistency.
            \item \textbf{If the rule cannot be naturally applied, you may skip the transformation.} Apply rules for naturalness, not for every applicable candidate.
        \end{itemize}

        \vspace{0.8em}
        \textbf{Format your response as:} \\
        \texttt{CONVERTED DIALOGUE: [Provide the \{target\_dialect\} version here]} \\
        \texttt{EXPLANATION OF CHANGES: [Provide your explanation here]}

        \vspace{0.8em}
        \textbf{\#\# Dialect Rule} \\
        \textbf{Rule:} \texttt{\{transformation\_rule\}} \\
        \textbf{Example of original sentence:} \texttt{\{original\_sentence\}} \\
        \textbf{Example of transformed sentence:} \texttt{\{transformed\_sentence\}} \\
        \textbf{Description:} \texttt{\{rule\_description\}} \\
        \texttt{Related Human's Comments: \{human\_comments\}}

        \vspace{0.8em}
        \textbf{\#\# New dialogue to transform (\{[user or assistant]\}):} \texttt{\{utterance\_to\_transform\}}
    \end{promptbox}

    \caption{Prompt for the construction of DialectLLM-User and DialectLLM-Model datasets.}
\end{figure}
\clearpage

\subsection{Prompt for Sec.~\ref{subsec:qc_revision}}
\label{app:qc_revision_prompt}
\begin{figure}[ht]
   \centering
    \begin{promptbox}{Wordbank Aggregation (filling out the missing words)}
        \sffamily
        Task: You are a linguist of \texttt{\{target\_dialect\}}  English. You will be given a Standard American \texttt{\{orthographic/lexical\}}  term. Output the \texttt{\{target\_dialect\}} English equivalent ONLY if it meets these criteria:

        \vspace{0.8em}
        \textbf{Rules:}
        \begin{itemize}[leftmargin=1.5em, nosep]
            \item It is a commonly used, mainstream \texttt{\{target\_dialect\}}  English term (not rare, archaic, or regional)
            \item It is the standard/primary term used in \texttt{\{target\_dialect\}} English
            \item \textbf{If the Standard American English term is used mostly or if no such prevalent \texttt{\{target\_dialect\}} equivalent exists, output the Standard American English \texttt{\{orthographic/lexical\}} term unchanged.}
        \end{itemize}

        \vspace{0.8em}
        The Standard American English \texttt{\{orthographic/lexical\}} term to translate: \\
        \texttt{\{us\_word\}} \\
        \vspace{0.8em}
        \textbf{Format:}

        \texttt{WORD: <the \texttt{\{target\_dialect\}} English equivalent or the original US term>}
    \end{promptbox}

    \caption{Prompt for filling out the missing words in the wordbank aggregation step. (See Fig.~\ref{fig:wb_filling} for visual depiction)}
\end{figure}
\begin{figure}[ht]
   \centering
    \begin{promptbox}{Wordbank Aggregation (filling out the missing ratings)}
        \sffamily
        Task: Compare the original (US English) word/spelling with the transformed (\texttt{\{target\_dialect\}} English) variant on a 1-4 scale:

        \vspace{0.8em}
        \textbf{Rules:}
        \begin{itemize}[leftmargin=1.5em, nosep]
            \item \textbf{1:} Feature is not used. The transformed word/spelling is never correct in \texttt{\{target\_dialect\}} English.
            \item \textbf{2:} Feature exists but is rare. The word/spelling might be understood but is outdated, region-specific, or influenced by other dialects. LLM should understand this term but not generate it.
            \item \textbf{3:} Feature exists and is commonly used in colloquial speech. The word/spelling is widespread in informal \texttt{\{target\_dialect\}} English but may coexist with other variants. Both forms are acceptable.
            \item \textbf{4:} Feature is required in this locale. Using the US variant would sound unnatural or incorrect. The \texttt{\{target\_dialect\}} variant is the standard and expected form.
        \end{itemize}

        \vspace{0.8em}
        You will be given a American English version of a word/spelling and its corresponding \texttt{\{target\_dialect\}} English transformation. Based on the criteria above, provide a rating from 1 to 4.
        
        \vspace{0.8em}
        \textbf{Reference Dataset:} \\
        \texttt{\{human annotated guideline\}}

        \vspace{0.8em}
        \textbf{Words to Evaluate:} \\
        US English: \texttt{\{us\_word\}} \\
        \texttt{\{target\_dialect\}} English: \texttt{\{target\_dialect\_word\}}

        \vspace{0.8em}
        \textbf{Format:} \\
        \texttt{Rating: <number from 1 to 4>}
    \end{promptbox}

    \caption{Prompt for filling out the missing ratings in the wordbank aggregation step. (See Fig.~\ref{fig:wb_filling} for visual depiction)}
\end{figure}
\begin{figure}[ht]
    \centering
    \begin{promptbox}{Australian English OrthoLex Refinement Guideline}
        \sffamily
        \textbf{en-AU General Orthographic (Spelling) Conventions}
        \begin{itemize}[leftmargin=1.5em, nosep]
            \item -our (e.g., colour, flavour, honour)
            \item -re (e.g., centre, theatre, metre)
            \item -ise (e.g., organise, realise)
            \item -yse (e.g., analyse, paralyse) 
            \item -ce (e.g., defence, licence, offence)
            \item -ogue (e.g., catalogue, dialogue) 
            \item e (e.g., encyclopedia, maneuver) OR ae and oe (e.g., encyclopaedia, manoeuvre) both acceptable
            \item doubled consonant (e.g., travelling, jewellery)
        \end{itemize}

        \vspace{1em}
        
        \textbf{en-AU Specific Lexical \& Orthographic Transformations}
        \vspace{0.5em}
        
        \texttt{| Original | Transformed | Category |} \\
        \texttt{| aubergine | eggplant | lexical |} \\
        \texttt{| yes | nah yeah (for responsive phrases only; other cases just stick with yes| lexical |}\\
        \texttt{| tire (wheel) | tyre (wheel) | orthographic |}
    \end{promptbox}

    \caption{An exemplar guideline for the OrthoLex refinement step for Australian English. We provide the model with general spelling conventions for all transformations, while specific changes are provided through probabilistic sampling when the terms appear in the dialogue (``aubergine'', ``yes'', ``tire'' in this example). These transformations are presented in a tabular format, as LLMs demonstrate superior performance when given information in structured forms~\cite{oh2024better}.}
\end{figure}

\clearpage
\subsection{Prompt for Sec.~\ref{subsec:exp_classification} }
\label{app:prompt_identification}
\begin{figure}[ht]
    \centering
    \begin{promptbox}{Dialect Identification (Classification) Prompt}
        \sffamily
        You are an expert linguist specializing in identifying different English dialects.

        \vspace{0.8em}
        Your task is to analyze a \texttt{\{turn\}}-turn dialogue where the user's and assistant's response is provided in a certain English dialect.

        \vspace{0.8em}
        The answer options are as follows:
        
        \begin{itemize}[label={}, leftmargin=0pt, nosep]
            \item (A) Indian English (India)
            \item (B) Canadian English (Canada)
            \item (C) Scottish English (Scotland)
            \item (D) Standard American English (USA)
            \item (E) Nigerian English (Nigeria)
            \item (F) Irish English (Ireland)
            \item (G) Australian English (Australia)
            \item (H) Philippine English (Philippine)
            \item (I) Standard British English (England)
        \end{itemize}

        \vspace{0.5em}
        \textbf{Instructions:} \\
        \begin{itemize}[leftmargin=1.5em, nosep]
            \item Carefully examine the morphosyntactic, lexical, and orthographic features in the dialogue.
            \item Based on your analysis, determine the dialect of the conversation.
            \item Provide your answer in the format specified below.
            \item Your answer should be one of the options above.
            \item Do not give explanations.
        \end{itemize}

        \vspace{0.8em}
        \textbf{Required Output Format:} \\
        \texttt{ANSWER: <Your Answer; Choose the correct option.>}
        
        \vspace{0.8em}
        \textbf{Dialogue to Evaluate:} \\
        \texttt{\{dialogue\}}
    \end{promptbox}

    \caption{Prompt template for dialect \benchmarkName classification task. Note that the answer options are shuffled for each instance and when testing the DialectLLM-OL dataset, we remove the mention of ``morphosyntactic'' for the first instruction.}
\end{figure}
\clearpage
\subsection{Prompt for Sec.~\ref{subsec:exp_response_completion}}
\label{app:prompt_response_completion}
\begin{figure}[ht]
    \centering
    \begin{promptbox}{Dialect Response Completion (Generation) Prompt}
        \sffamily
        You are an expert linguist specializing in \texttt{\{target\_dialect\}} English dialect.

        \vspace{0.8em}
        Your task is to select the most appropriate response to complete a \texttt{\{turn\}}-turn dialogue. The dialogue exhibits specific dialectal features: lexical, orthographic, and morphosyntactic features.

        \vspace{0.8em}
        You will be provided with:
        \begin{enumerate}[label=\arabic*., nosep, leftmargin=2em]
            \item The Conversation History (Conversation history of previous turns).
            \item The Current User Turn (The message the Assistant must reply to).
            \item A list of Candidate Options for the Assistant's response.
        \end{enumerate}

        \vspace{0.8em}
        \textbf{Instructions:}
        \begin{itemize}[leftmargin=1.5em, nosep]
            \item Analyze the \texttt{\{sentence\}} of the 'Conversation History' and 'Current User Turn' to identify the specific English dialect being used.
            \item Evaluate the 'Candidate Options' and select the one that is linguistically consistent with the identified dialect from the conversation history and current user turn.
            \item Provide your answer in the format specified below.
            \item Your answer should be one of the options above (provide the corresponding index/label). Choose the most probable option.
            \item Do not give explanations.
        \end{itemize}

        \vspace{0.8em}
        \textbf{Required Output Format:} \\
        \texttt{ANSWER: <Your Answer; Choose the correct option; Provide the alphabet character of the chosen option.>}

        \vspace{0.8em}
        \textbf{Conversation History:} \\
        \texttt{\{context\_history\}}

        \vspace{0.5em}
        \textbf{Current User Turn:} \\
        \texttt{\{current\_user\_turn\}}

        \vspace{0.5em}
        \textbf{Candidate Options:} \\
        \texttt{\{possible\_options\_str\}}
    \end{promptbox}

    \caption{Prompt template for dialect \benchmarkName generation task. Note that the answer options are shuffled for each instance and when testing the DialectLLM-OL dataset, we remove the mention of ``morphosyntactic'' features.}
\end{figure}
\clearpage
\subsection{Prompt for Appendix~\ref{app:llm_generation}}
\begin{figure}[ht]
    \centering
    \begin{promptbox}{LLM-as-a-Judge (Lexical Judge) for Open-Ended Generation Evaluation}
        \sffamily
        You are an expert linguist specializing in English lexical and vocabulary variations for \texttt{\{target\_dialect]\}} English dialect. Your task is to evaluate a text based STRICTLY on its lexical choices (vocabulary) for \texttt{\{target\_dialect]\}} English dialect.

        \vspace{0.8em}
        \textbf{Evaluation Guidelines:} \\
        \texttt{\{user\_provided\_guidelines\}}

        \vspace{0.8em}
        \textbf{Binary Classification Scale:} \\
        \textbf{1 - Appropriate} (ALL vocabulary choices are consistent with \texttt{\{target\_dialect]\}} English, or the text uses universal/neutral terms that do not violate dialect norms). \\
        \textbf{0 - Inappropriate} (The text contains ONE OR MORE words or idioms that belong to a conflicting dialect). \\
        \textbf{N/A - Not Applicable} (The text contains no words that have variant lexical aspects/vocabularies across dialects).

        \vspace{0.8em}
        \textbf{Instructions:}
        \begin{itemize}[leftmargin=1.5em, nosep]
            \item The 'Evaluation Guidelines' consist of the human annotation guidelines and their corresponding annotations.
            \item Use the 'Evaluation Guidelines' as a reference frame to identify acceptable variations (STRICTLY FOLLOW THE GUIDELINES GIVEN).
            \item \textbf{SCORE CRITERIA:} To receive a score of 1, ``ALL'' lexical components must be appropriate. If the text uses a term from a conflicting dialect or is inappropriate, the score must be 0. If the input text does not contain any words that differ in vocabulary between dialects, output ``N/A''.
            \item Focus ONLY on word/vocabulary choices.
            \item Ignore spelling and grammatical structure.
            \item Please detail the factors contributing to this assessment.
        \end{itemize}

        \vspace{0.8em}
        \textbf{Required Output Format:} \\
        \texttt{Reasoning: <Explanation>} \\
        \texttt{Score: <0, 1, or N/A>}

        \vspace{0.8em}
        \textbf{Input Text:} \\
        \texttt{\{response\_to\_evaluate\}}
    \end{promptbox}
    \caption{Prompt for LLMaaJ that evaluates lexical feature appropriateness.}
\end{figure}
%%%%%%%%%%%%%%%%%%%%%%%%%%%%%%%%%%%%%%%%%%%%%%%%%%%%%%%%%%%%%%%%%%%%%%%%%%%%%%%
%%%%%%%%%%%%%%%%%%%%%%%%%%%%%%%%%%%%%%%%%%%%%%%%%%%%%%%%%%%%%%%%%%%%%%%%%%%%%%%
\newpage
\section*{NeurIPS Paper Checklist}

\begin{enumerate}

\item {\bf Claims}
    \item[] Question: Do the main claims made in the abstract and introduction accurately reflect the paper's contributions and scope?
    \item[] Answer: \answerYes{} % Replace by \answerYes{}, \answerNo{}, or \answerNA{}.
    \item[] Justification: We clearly state our contributions and scope in both abstract and introduction.
    \item[] Guidelines:
    \begin{itemize}
        \item The answer \answerNA{} means that the abstract and introduction do not include the claims made in the paper.
        \item The abstract and/or introduction should clearly state the claims made, including the contributions made in the paper and important assumptions and limitations. A \answerNo{} or \answerNA{} answer to this question will not be perceived well by the reviewers. 
        \item The claims made should match theoretical and experimental results, and reflect how much the results can be expected to generalize to other settings. 
        \item It is fine to include aspirational goals as motivation as long as it is clear that these goals are not attained by the paper. 
    \end{itemize}

\item {\bf Limitations}
    \item[] Question: Does the paper discuss the limitations of the work performed by the authors?
    \item[] Answer: \answerYes{} % Replace by \answerYes{}, \answerNo{}, or \answerNA{}.
    \item[] Justification: Limitations are discussed in Section \ref{sec:conclusion}.
    \item[] Guidelines:
    \begin{itemize}
        \item The answer \answerNA{} means that the paper has no limitation while the answer \answerNo{} means that the paper has limitations, but those are not discussed in the paper. 
        \item The authors are encouraged to create a separate ``Limitations'' section in their paper.
        \item The paper should point out any strong assumptions and how robust the results are to violations of these assumptions (e.g., independence assumptions, noiseless settings, model well-specification, asymptotic approximations only holding locally). The authors should reflect on how these assumptions might be violated in practice and what the implications would be.
        \item The authors should reflect on the scope of the claims made, e.g., if the approach was only tested on a few datasets or with a few runs. In general, empirical results often depend on implicit assumptions, which should be articulated.
        \item The authors should reflect on the factors that influence the performance of the approach. For example, a facial recognition algorithm may perform poorly when image resolution is low or images are taken in low lighting. Or a speech-to-text system might not be used reliably to provide closed captions for online lectures because it fails to handle technical jargon.
        \item The authors should discuss the computational efficiency of the proposed algorithms and how they scale with dataset size.
        \item If applicable, the authors should discuss possible limitations of their approach to address problems of privacy and fairness.
        \item While the authors might fear that complete honesty about limitations might be used by reviewers as grounds for rejection, a worse outcome might be that reviewers discover limitations that aren't acknowledged in the paper. The authors should use their best judgment and recognize that individual actions in favor of transparency play an important role in developing norms that preserve the integrity of the community. Reviewers will be specifically instructed to not penalize honesty concerning limitations.
    \end{itemize}

\item {\bf Theory assumptions and proofs}
    \item[] Question: For each theoretical result, does the paper provide the full set of assumptions and a complete (and correct) proof?
    \item[] Answer: \answerNA{} % Replace by \answerYes{}, \answerNo{}, or \answerNA{}.
    \item[] Justification: No theoretical results exist in the paper.
    \item[] Guidelines:
    \begin{itemize}
        \item The answer \answerNA{} means that the paper does not include theoretical results. 
        \item All the theorems, formulas, and proofs in the paper should be numbered and cross-referenced.
        \item All assumptions should be clearly stated or referenced in the statement of any theorems.
        \item The proofs can either appear in the main paper or the supplemental material, but if they appear in the supplemental material, the authors are encouraged to provide a short proof sketch to provide intuition. 
        \item Inversely, any informal proof provided in the core of the paper should be complemented by formal proofs provided in appendix or supplemental material.
        \item Theorems and Lemmas that the proof relies upon should be properly referenced. 
    \end{itemize}

    \item {\bf Experimental result reproducibility}
    \item[] Question: Does the paper fully disclose all the information needed to reproduce the main experimental results of the paper to the extent that it affects the main claims and/or conclusions of the paper (regardless of whether the code and data are provided or not)?
    \item[] Answer: \answerYes{}% Replace by \answerYes{}, \answerNo{}, or \answerNA{}.
    \item[] Justification: The whole framework and experimental details are presented in the main paper (Section \ref{sec:data_generation}, \ref{sec:dialectllm-bench}, and \ref{sec:results}) and the corresponding prompts are also provided in Appendix \ref{app:prompts}.
    \item[] Guidelines:
    \begin{itemize}
        \item The answer \answerNA{} means that the paper does not include experiments.
        \item If the paper includes experiments, a \answerNo{} answer to this question will not be perceived well by the reviewers: Making the paper reproducible is important, regardless of whether the code and data are provided or not.
        \item If the contribution is a dataset and\slash or model, the authors should describe the steps taken to make their results reproducible or verifiable. 
        \item Depending on the contribution, reproducibility can be accomplished in various ways. For example, if the contribution is a novel architecture, describing the architecture fully might suffice, or if the contribution is a specific model and empirical evaluation, it may be necessary to either make it possible for others to replicate the model with the same dataset, or provide access to the model. In general. releasing code and data is often one good way to accomplish this, but reproducibility can also be provided via detailed instructions for how to replicate the results, access to a hosted model (e.g., in the case of a large language model), releasing of a model checkpoint, or other means that are appropriate to the research performed.
        \item While NeurIPS does not require releasing code, the conference does require all submissions to provide some reasonable avenue for reproducibility, which may depend on the nature of the contribution. For example
        \begin{enumerate}
            \item If the contribution is primarily a new algorithm, the paper should make it clear how to reproduce that algorithm.
            \item If the contribution is primarily a new model architecture, the paper should describe the architecture clearly and fully.
            \item If the contribution is a new model (e.g., a large language model), then there should either be a way to access this model for reproducing the results or a way to reproduce the model (e.g., with an open-source dataset or instructions for how to construct the dataset).
            \item We recognize that reproducibility may be tricky in some cases, in which case authors are welcome to describe the particular way they provide for reproducibility. In the case of closed-source models, it may be that access to the model is limited in some way (e.g., to registered users), but it should be possible for other researchers to have some path to reproducing or verifying the results.
        \end{enumerate}
    \end{itemize}

\item {\bf Open access to data and code}
    \item[] Question: Does the paper provide open access to the data and code, with sufficient instructions to faithfully reproduce the main experimental results, as described in supplemental material?
    \item[] Answer: \answerNo{} % Replace by \answerYes{}, \answerNo{}, or \answerNA{}.
    \item[] Justification: Currently, we are going through the approval processes required to release the data and code as part of this research in accordance with our organization’s guidelines, and our ability to release the artifacts is predicated on the approval decision. We have provided detailed steps on dataset generation (Section \ref{sec:data_generation}), along with the prompts used (Appendix \ref{app:prompts}). After approval, the data and code will be released on our project page (\url{https://harryoh99.github.io/DialectLLM/}).
    \item[] Guidelines:
    \begin{itemize}
        \item The answer \answerNA{} means that paper does not include experiments requiring code.
        \item Please see the NeurIPS code and data submission guidelines (\url{https://neurips.cc/public/guides/CodeSubmissionPolicy}) for more details.
        \item While we encourage the release of code and data, we understand that this might not be possible, so \answerNo{} is an acceptable answer. Papers cannot be rejected simply for not including code, unless this is central to the contribution (e.g., for a new open-source benchmark).
        \item The instructions should contain the exact command and environment needed to run to reproduce the results. See the NeurIPS code and data submission guidelines (\url{https://neurips.cc/public/guides/CodeSubmissionPolicy}) for more details.
        \item The authors should provide instructions on data access and preparation, including how to access the raw data, preprocessed data, intermediate data, and generated data, etc.
        \item The authors should provide scripts to reproduce all experimental results for the new proposed method and baselines. If only a subset of experiments are reproducible, they should state which ones are omitted from the script and why.
        \item At submission time, to preserve anonymity, the authors should release anonymized versions (if applicable).
        \item Providing as much information as possible in supplemental material (appended to the paper) is recommended, but including URLs to data and code is permitted.
    \end{itemize}

\item {\bf Experimental setting/details}
    \item[] Question: Does the paper specify all the training and test details (e.g., data splits, hyperparameters, how they were chosen, type of optimizer) necessary to understand the results?
    \item[] Answer: \answerYes{} % Replace by \answerYes{}, \answerNo{}, or \answerNA{}.
    \item[] Justification: Experiment settings on evaluation and training are explained in Section \ref{sec:results} and Appendix \ref{app:model_hyperparameter} of the paper.
    \item[] Guidelines:
    \begin{itemize}
        \item The answer \answerNA{} means that the paper does not include experiments.
        \item The experimental setting should be presented in the core of the paper to a level of detail that is necessary to appreciate the results and make sense of them.
        \item The full details can be provided either with the code, in appendix, or as supplemental material.
    \end{itemize}

\item {\bf Experiment statistical significance}
    \item[] Question: Does the paper report error bars suitably and correctly defined or other appropriate information about the statistical significance of the experiments?
    \item[] Answer: \answerYes{} % Replace by \answerYes{}, \answerNo{}, or \answerNA{}.
    \item[] Justification: For experiments that result in numerical results, to control randomness, the experiments are proceeded with temperature 0, resulting in deterministic results as mentioned in Appendix \ref{app:hyperparameters}. We also report the 10K bootstrap uncertainty estimates for the main experiment in Sec. \ref{subsec:main_results}.
    \item[] Guidelines:
    \begin{itemize}
        \item The answer \answerNA{} means that the paper does not include experiments.
        \item The authors should answer \answerYes{} if the results are accompanied by error bars, confidence intervals, or statistical significance tests, at least for the experiments that support the main claims of the paper.
        \item The factors of variability that the error bars are capturing should be clearly stated (for example, train/test split, initialization, random drawing of some parameter, or overall run with given experimental conditions).
        \item The method for calculating the error bars should be explained (closed form formula, call to a library function, bootstrap, etc.)
        \item The assumptions made should be given (e.g., Normally distributed errors).
        \item It should be clear whether the error bar is the standard deviation or the standard error of the mean.
        \item It is OK to report 1-sigma error bars, but one should state it. The authors should preferably report a 2-sigma error bar than state that they have a 96\% CI, if the hypothesis of Normality of errors is not verified.
        \item For asymmetric distributions, the authors should be careful not to show in tables or figures symmetric error bars that would yield results that are out of range (e.g., negative error rates).
        \item If error bars are reported in tables or plots, the authors should explain in the text how they were calculated and reference the corresponding figures or tables in the text.
    \end{itemize}

\item {\bf Experiments compute resources}
    \item[] Question: For each experiment, does the paper provide sufficient information on the computer resources (type of compute workers, memory, time of execution) needed to reproduce the experiments?
    \item[] Answer: \answerYes{} % Replace by \answerYes{}, \answerNo{}, or \answerNA{}.
    \item[] Justification: The details on the resources are detailed in Appendix \ref{app:models_database_details}.
    \item[] Guidelines:
    \begin{itemize}
        \item The answer \answerNA{} means that the paper does not include experiments.
        \item The paper should indicate the type of compute workers CPU or GPU, internal cluster, or cloud provider, including relevant memory and storage.
        \item The paper should provide the amount of compute required for each of the individual experimental runs as well as estimate the total compute. 
        \item The paper should disclose whether the full research project required more compute than the experiments reported in the paper (e.g., preliminary or failed experiments that didn't make it into the paper). 
    \end{itemize}
    
\item {\bf Code of ethics}
    \item[] Question: Does the research conducted in the paper conform, in every respect, with the NeurIPS Code of Ethics \url{https://neurips.cc/public/EthicsGuidelines}?
    \item[] Answer: \answerYes{} % Replace by \answerYes{}, \answerNo{}, or \answerNA{}.
    \item[] Justification: The authors strictly follow the NeurIPS Code of Ethics.
    \item[] Guidelines:
    \begin{itemize}
        \item The answer \answerNA{} means that the authors have not reviewed the NeurIPS Code of Ethics.
        \item If the authors answer \answerNo, they should explain the special circumstances that require a deviation from the Code of Ethics.
        \item The authors should make sure to preserve anonymity (e.g., if there is a special consideration due to laws or regulations in their jurisdiction).
    \end{itemize}

\item {\bf Broader impacts}
    \item[] Question: Does the paper discuss both potential positive societal impacts and negative societal impacts of the work performed?
    \item[] Answer: \answerYes{} % Replace by \answerYes{}, \answerNo{}, or \answerNA{}.
    \item[] Justification: The potential (broader) impact is discussed in Appendix \ref{app:Broader_Impact}.
    \item[] Guidelines:
    \begin{itemize}
        \item The answer \answerNA{} means that there is no societal impact of the work performed.
        \item If the authors answer \answerNA{} or \answerNo, they should explain why their work has no societal impact or why the paper does not address societal impact.
        \item Examples of negative societal impacts include potential malicious or unintended uses (e.g., disinformation, generating fake profiles, surveillance), fairness considerations (e.g., deployment of technologies that could make decisions that unfairly impact specific groups), privacy considerations, and security considerations.
        \item The conference expects that many papers will be foundational research and not tied to particular applications, let alone deployments. However, if there is a direct path to any negative applications, the authors should point it out. For example, it is legitimate to point out that an improvement in the quality of generative models could be used to generate Deepfakes for disinformation. On the other hand, it is not needed to point out that a generic algorithm for optimizing neural networks could enable people to train models that generate Deepfakes faster.
        \item The authors should consider possible harms that could arise when the technology is being used as intended and functioning correctly, harms that could arise when the technology is being used as intended but gives incorrect results, and harms following from (intentional or unintentional) misuse of the technology.
        \item If there are negative societal impacts, the authors could also discuss possible mitigation strategies (e.g., gated release of models, providing defenses in addition to attacks, mechanisms for monitoring misuse, mechanisms to monitor how a system learns from feedback over time, improving the efficiency and accessibility of ML).
    \end{itemize}
    
\item {\bf Safeguards}
    \item[] Question: Does the paper describe safeguards that have been put in place for responsible release of data or models that have a high risk for misuse (e.g., pre-trained language models, image generators, or scraped datasets)?
    \item[] Answer: \answerNA{} % Replace by \answerYes{}, \answerNo{}, or \answerNA{}.
    \item[] Justification: The paper poses no such risks.
    \item[] Guidelines:
    \begin{itemize}
        \item The answer \answerNA{} means that the paper poses no such risks.
        \item Released models that have a high risk for misuse or dual-use should be released with necessary safeguards to allow for controlled use of the model, for example by requiring that users adhere to usage guidelines or restrictions to access the model or implementing safety filters. 
        \item Datasets that have been scraped from the Internet could pose safety risks. The authors should describe how they avoided releasing unsafe images.
        \item We recognize that providing effective safeguards is challenging, and many papers do not require this, but we encourage authors to take this into account and make a best faith effort.
    \end{itemize}

\item {\bf Licenses for existing assets}
    \item[] Question: Are the creators or original owners of assets (e.g., code, data, models), used in the paper, properly credited and are the license and terms of use explicitly mentioned and properly respected?
    \item[] Answer: \answerYes{} % Replace by \answerYes{}, \answerNo{}, or \answerNA{}.
    \item[] Justification: The paper generates data with LLMs, while the licenses of existing resource used are detailed in Appendix \ref{app:models_database_details}. 
    \item[] Guidelines:
    \begin{itemize}
        \item The answer \answerNA{} means that the paper does not use existing assets.
        \item The authors should cite the original paper that produced the code package or dataset.
        \item The authors should state which version of the asset is used and, if possible, include a URL.
        \item The name of the license (e.g., CC-BY 4.0) should be included for each asset.
        \item For scraped data from a particular source (e.g., website), the copyright and terms of service of that source should be provided.
        \item If assets are released, the license, copyright information, and terms of use in the package should be provided. For popular datasets, \url{paperswithcode.com/datasets} has curated licenses for some datasets. Their licensing guide can help determine the license of a dataset.
        \item For existing datasets that are re-packaged, both the original license and the license of the derived asset (if it has changed) should be provided.
        \item If this information is not available online, the authors are encouraged to reach out to the asset's creators.
    \end{itemize}

\item {\bf New assets}
    \item[] Question: Are new assets introduced in the paper well documented and is the documentation provided alongside the assets?
    \item[] Answer: \answerNA{} % Replace by \answerYes{}, \answerNo{}, or \answerNA{}.
    \item[] Justification: This submission focuses on the framework and does not include the release of any new assets. The details of the framework is detailed in the paper.
    \item[] Guidelines:
    \begin{itemize}
        \item The answer \answerNA{} means that the paper does not release new assets.
        \item Researchers should communicate the details of the dataset\slash code\slash model as part of their submissions via structured templates. This includes details about training, license, limitations, etc. 
        \item The paper should discuss whether and how consent was obtained from people whose asset is used.
        \item At submission time, remember to anonymize your assets (if applicable). You can either create an anonymized URL or include an anonymized zip file.
    \end{itemize}

\item {\bf Crowdsourcing and research with human subjects}
    \item[] Question: For crowdsourcing experiments and research with human subjects, does the paper include the full text of instructions given to participants and screenshots, if applicable, as well as details about compensation (if any)? 
    \item[] Answer: \answerYes{} % Replace by \answerYes{}, \answerNo{}, or \answerNA{}.
    \item[] Justification: We include the instruction templates given to users for human studies in Appendix \ref{app:human_annotation_instruction}, \ref{app:preference_annotations}, and \ref{app:user_model_human_detail}. Details of human annotators are described in Appendix \ref{subsec:annot_recruitment}.
    \item[] Guidelines:
    \begin{itemize}
        \item The answer \answerNA{} means that the paper does not involve crowdsourcing nor research with human subjects.
        \item Including this information in the supplemental material is fine, but if the main contribution of the paper involves human subjects, then as much detail as possible should be included in the main paper. 
        \item According to the NeurIPS Code of Ethics, workers involved in data collection, curation, or other labor should be paid at least the minimum wage in the country of the data collector. 
    \end{itemize}

\item {\bf Institutional review board (IRB) approvals or equivalent for research with human subjects}
    \item[] Question: Does the paper describe potential risks incurred by study participants, whether such risks were disclosed to the subjects, and whether Institutional Review Board (IRB) approvals (or an equivalent approval/review based on the requirements of your country or institution) were obtained?
    \item[] Answer: \answerNo{} % Replace by \answerYes{}, \answerNo{}, or \answerNA{}.
    \item[] Justification: This work involves human annotation performed by hired resources as part of their standard job responsibilities. Annotators provided labels on data samples and were not themselves the subject of analysis. The task did not involve collection of personal or sensitive information about annotators, and no identifiable data was recorded. The annotation process is considered minimal/no risk under research guidelines. Based on these factors, formal IRB approval or equivalent external review was not required. Appropriate compliance and data handling standards were followed.
    \item[] Guidelines:
    \begin{itemize}
        \item The answer \answerNA{} means that the paper does not involve crowdsourcing nor research with human subjects.
        \item Depending on the country in which research is conducted, IRB approval (or equivalent) may be required for any human subjects research. If you obtained IRB approval, you should clearly state this in the paper. 
        \item We recognize that the procedures for this may vary significantly between institutions and locations, and we expect authors to adhere to the NeurIPS Code of Ethics and the guidelines for their institution. 
        \item For initial submissions, do not include any information that would break anonymity (if applicable), such as the institution conducting the review.
    \end{itemize}

\item {\bf Declaration of LLM usage}
    \item[] Question: Does the paper describe the usage of LLMs if it is an important, original, or non-standard component of the core methods in this research? Note that if the LLM is used only for writing, editing, or formatting purposes and does \emph{not} impact the core methodology, scientific rigor, or originality of the research, declaration is not required.
    %this research? 
    \item[] Answer: \answerYes{} % Replace by \answerYes{}, \answerNo{}, or \answerNA{}.
    \item[] Justification: DialectLLM incorporates an LLM (Claude-4-Sonnet) as a tool for generating the proposed dataset. These points are mentioned repeatedly in the paper in detail. Besides this point, the core method development in this research does not involve LLMs as any important, original, or non-standard components.
    \item[] Guidelines:
    \begin{itemize}
        \item The answer \answerNA{} means that the core method development in this research does not involve LLMs as any important, original, or non-standard components.
        \item Please refer to our LLM policy in the NeurIPS handbook for what should or should not be described.
    \end{itemize}

\end{enumerate}
\end{document}